\documentclass[11pt]{article}

\usepackage[a4paper, total={6in, 9in}]{geometry}

\usepackage{amssymb}
\usepackage{amsmath}

\usepackage{multirow}

\usepackage{todonotes}
\usepackage{rotating}

\usepackage{amsthm}
\usepackage{natbib}

\usepackage{graphicx}
\usepackage{makecell}
\usepackage{array}
\newcolumntype{M}[1]{>{\centering\arraybackslash}m{#1}}

\newtheorem{example}{Example}
\newtheorem{theorem}{Theorem}
\newtheorem{lemma}{Lemma}
\newtheorem{definition}{Definition}

\usepackage{algorithm}
\usepackage{algpseudocode}

\newcommand{\algorithmicbreak}{\textbf{break}}
\newcommand{\BREAK}{\State \algorithmicbreak}

\usepackage{subcaption}

\usepackage{booktabs}

\usepackage{makecell}
\usepackage{array}
\newcolumntype{M}[1]{>{\centering\arraybackslash}m{#1}}

\title{Data as Voters: Core Set Selection Using Approval-Based Multi-Winner
  Voting}

\author{Luis S\'anchez-Fern\'andez\footnote{e-mail: luiss@it.uc3m.es}, Jes\'us A. Fisteus\footnote{jaf@it.uc3m.es},\\ Rafael Eugenio L\'opez Zaragoza\\Dept. Telematic Engineering, Universidad Carlos III de Madrid, \\Av. Universidad, 30, Legan\'es (Madrid), E-28911, Madrid, Spain}

\providecommand{\keywords}[1]
{
  \small	
  \textbf{\textit{Keywords---}} #1
}

\begin{document}

\maketitle

\begin{abstract}
We present a novel approach to the core set/instance selection problem in machine learning.
Our approach is based on recent results on (proportional)
representation in approval-based multi-winner elections. In our model,
instances play a double role as voters and candidates. The approval set
of each instance in the training set (acting as a voter) is defined from
the concept of local set, which already exists in the literature. We then select the
election winners by using a representative voting rule, and such
winners are the data instances kept in the reduced training set.
We evaluate our approach in two experiments involving neural network classifiers
and classic machine learning classifiers (KNN and SVM). Our experiments show that,
in several cases, our approach improves the performance of state-of-the-art
methods, and the differences are statistically significant.
\end{abstract}


\section{Introduction}
\label{sec:intro}

Efficiency in the use of computing resources has been a key driving factor in the development of machine learning techniques from their inception in the $1950$s. This search for efficiency naturally justifies the development of techniques to obtain a reduced dataset from a larger original dataset that could be used as input for a machine learning algorithm, while keeping the loss in performance compared to the original dataset as low as possible. Initial work, such as {\it Condensed Nearest Neighbour} (CNN), proposed by~\citet{hart1968condensed}, were motivated by the computational requirements of the {\it K-Nearest Neighbours} (KNN) classifier\footnote{A naive algorithm to classify a new instance with a KNN classifier requires storing the entire training set and computing the distance between the new instance and each instance in the training set. Therefore, the time needed to classify a new instance grows linearly with the size of the training set. It can be considerably reduced with later approaches based on organising the training instances in specific data structures such as K-D trees~\citep{tiwari2023developments}. However, it still grows (sublinearly) with the size of the dataset.} and the limited computing resources available at that time. A different motivation is behind another pioneering approach: The {\it Edited Nearest Neighbour}, proposed by~\citet{wilson1972asymptotic}. The focus of ENN is to remove noisy instances from the original dataset. It obtains a modest reduction, but often improves the performance of the original dataset.
In classic machine learning, the problem of obtaining a reduced dataset from a larger dataset was referred to as {\it instance selection}~\citep{garcia2015data}.

In the Big Data era~\citep{lv2017next}, storing, preprocessing, and training large datasets requires costly computing resources. Thus, techniques that allow for the reduction of the size of the data employed in AI-driven applications can dramatically affect their production cost.\footnote{An example of the influence of the data size on the costs of AI tools is the appearance of the DeepSeek chatbot in January 2025. DeepSeek chatbot appeared as a competitor of OpenAI ChatGPT, with training costs significantly lower than those of other LLMs. OpenAI accused them of using distilled (a technique that we will discuss later) data from OpenAI to reduce costs. https://www.binance.com/en/square/post/19592530706497 (visited on 6/6/2025).} A common situation is that models are not trained only once because new data is being received continuously. An example of this type of scenario is {\it class incremental learning}~\citep{rebuffi2017icarl,wang2025class}, where we have a stream that, occasionally, receives a new class together with a set of instances that belong to it. We need to build a classifier that can recognise all the classes seen so far at certain time points. Incremental training with stochastic gradient descent optimisation using only new instances whenever they arrive is not possible due to the problem of {\it catastrophic forgetting}~\citep{mccloskey1989catastrophic}.
Storing all the instances that have been received is neither possible because of its cost. Therefore, it is necessary to select a subset of the instances of each received class that will be stored.

Motivated by the computational costs of large datasets, the topic of obtaining a reduced dataset from a large original one has recently been rediscovered in deep learning and other machine learning applications~\citep{rebuffi2017icarl,cui2022dc}. The different approaches proposed can be classified into two groups: {\it core set selection}~\citep{rebuffi2017icarl,sener2018active} and {\it dataset condensation}\footnote{The term `condensation' in classic instance selection has a different meaning, as we will discuss in the Related Work.} (or {\it dataset distillation})~\citep{10275116}. The difference between core set selection and dataset condensation is that the goal of the former is to select a subset of the instances in the original dataset while the goal of the latter is to produce a small set of synthesised instances that are ``summaries'' of the instances in the original dataset.\footnote{Proposals for producing synthesised samples from the original dataset in classic machine learning were referred to as {\it prototype generation}~\citep{triguero2011taxonomy}.}

This paper focuses on core set/instance selection (in what follows, we will refer to core set/instance selection as CS/I selection). Our approach is based on the following observation. Some previous work in the state of the art~\citep{rebuffi2017icarl,castro2018end,zhao2021dataset2,zhang2025compr} propose approaches that aim to find a set of (real or synthetic) instances that are {\it representative} of the original dataset. In a completely different setting, the computational social choice~\citep{brandt2016handbook} community has extensively studied the concept of representation in approval-based multi-winner voting~\citep{multiwinnerApprovalSurvey}.
Our approach is to exploit recent results in that field to build a solution for the CS/I selection problem.

In our model, instances play a double role as voters and candidates.
Specifically, each instance has associated a certain area of interest and, acting as a voter, approves of instances that are placed in such area.
Then, using notions of proportionality taken from the computational social choice domain, we select a subset of the instances that are representative of the region occupied by each class.
The area of interest of each instance is identified using the notion of local set.
The local set of an instance, introduced first by~\citet{brighton2002advances}, is the set of instances of the same class closer to it than the nearest instance of a different class.
These ideas will be further ellaborated in Section~\ref{sec:model}.

{\bf Our Contribution:} We first formalise the notion of approval-based multi-winner election that is local-set-derived from a CS/I selection problem. Then, we prove a theoretical result that guarantees that any instance in the original training set that approves of at least $(\textrm{K}+1)/2$ instances (assuming that K is odd)  will be correctly classified with a KNN classifier trained with the reduced training set obtained with representative voting rules that satisfy certain conditions. Then, we show graphically the effect of reducing a training set using representative voting rules
with Principal Component Analysis (PCA) techniques. We then evaluate our approach with two different experiments. In the first one, we train several neural networks with reduced datasets and compare our approach with several algorithms proposed in the state of the art for core set selection. Our results show that the reduced datasets obtained with our approach outperform all the state-of-the-art algorithms in accuracy, except in one case, and differences are statistically significant.
In the second experiment, we evaluate our approach with a KNN and a SVM classifier, and we compare our approach with several instance selection algorithms proposed in the state of the art.
We also validate our approach with a statistical evaluation of the results of this experiment.

\section{Related Work}
\label{sec:soa}

\subsection{Core set selection}

In this section, we introduce several core set selection methods from the state of the art. Herding, based on~\citep{welling2009herding}, consists of iteratively selecting the sample that makes the centre of the reduced set of samples of a given class as close as possible to the centre of the samples of such class in the original dataset. It has been used
in~\citep{rebuffi2017icarl,castro2018end,wu2019large,wu2025feature}.

$k$-Center selects the $k$ instances for each class that minimise the maximum distance between an instance of such class in the original dataset and its closest instance in the selected set of $k$ instances. $k$-Center is known to be NP-hard to compute, so it is usually approximated using heuristics. It has been used in~\citep{sener2018active}.

$k$-Means is a clustering algorithm. Given a set of samples, it groups them in $k$ clusters $\{X_1, ..., X_k\}$ to minimise $\sum_{i=1}^k \sum_{s \in X_i} ||s - \mu_i||^2$, where $\mu_i$ is the centre of cluster $X_i$. As in the case of $k$-Center, $k$-Means is NP-hard to compute, and is usually approximated using heuristics. Since $k$-Means is a clustering algorithm, it does not directly return a set of $k$ instances. After computing the clusters, the instance of each cluster closest to the cluster centre is selected. $k$-Means has been used in~\citep{cui2022dc}.\footnote{\citet{cui2022dc} say in the main paper that they use $k$-Center, but in the appendix they explain that they have actually used the $k$-Means implementation of scikit-learn. Since $k$-Means is substantially different from $k$-Center, they have used $k$-Means. \citet{cui2022dc} report substantially different (better) accuracy values for $k$-Center from those reported in other work (including this paper), which confirms that they have used $k$-Means and not $k$-Center in their experiments.}

\subsection{Instance selection}

Instance selection methods are typically classified in {\it condensation},\footnote{We recall that `condensation' has a different meaning in modern dataset reduction techniques.} {\it edition} and {\it hybrid} methods~\citep{garcia2015data}. Condensation methods try to keep the instances on the boundaries of each class and to remove inner instances; edition methods try to remove noisy instances (which are often on the boundaries); finally, hybrid methods typically apply an edition method first, to remove noisy instances, and then try to remove inner instances. In this section, we review some prominent approaches in the literature.

\subsubsection{Condensation methods}

{\it Condensed Nearest Neighbour} (CNN)~\citep{hart1968condensed}
was the first condensation method proposed. Its goal is to find a subset of the original dataset that can correctly classify all the instances in the original dataset with a $1$NN classifier. CNN maintains two data stores where the instances in the training set are allocated. Let us refer to them as TR and TEMP. Initially, TR contains one data instance, and TEMP contains all the remaining instances in the training set. CNN repeatedly traverses the instances in TEMP. For each one, its class is computed using the data in TR. If that class is correct, then the instance is kept in TEMP. Otherwise, it is moved from TEMP to TR. The process stops when no instance is moved from TEMP to TR at some iteration point.

{\it Local Density Instance Selection} (LDIS)~\citep{carbonera2015density} aims to select locally representative instances of each class in the training set. LDIS starts computing the {\it density} of each instance, which is defined as the opposite of the average distance between that instance and all the remaining instances of the same class in the training set. The notion of distance can be freely chosen.
The density of each instance is then compared to the densities of its closest $k$ instances of the same class (where $k$ is a parameter that can be freely chosen). The instances whose densities are greater than those of their $k$ closest neighbours of the same class are kept in the reduced training set.

{\it Instance Selection based on Dense
Spatial Partitions} (ISDSP)~\citep{carbonera2018efficient} first uses a parameter $p\in [0,1]$ to determine, for each class, the number $k$ of instances of such class that must be kept in the reduced training set, as the number of instances of such class in the original training set multiplied by $p$. Secondly, it determines, for each class and feature, the range of values that the instances in the training set belonging to the class can take for that feature. Those ranges are divided into $n$ equal-size intervals, where $n$ is a positive integer. This divides the space where the instances of each class are located in $nm$ equal-size $m$-dimensional intervals.
To determine the instances of each class that must be kept in the reduced training set, we first select the $k$ $m$-dimensional intervals that contain more instances of that class. Then, for each selected interval, the instance closer to the centroid of the instances of the given class included in it is selected.

\subsubsection{Edition methods}

The first edition method proposed was {\it Edited Nearest Neighbour} (ENN)~\citep{wilson1972asymptotic}, which is still used often as the first step in hybrid methods. ENN removes from the training set those instances whose class does not agree with the class of the majority of their $K$ nearest neighbours.

Later,
{\it Local Set-based Smoother} (LSSm)~\citep{leyva2015three}
was proposed.
LSSm uses the concepts of local set and nearest enemy to define two metrics for each instance: {\em usefulness} and {\em harmfulness}. The usefulness of an instance $d$ is the number of instances in the training set that have $d$ in their local sets. Its harmfulness is the number of instances in the training set such that $d$ is their nearest enemy. LSSm removes from the training set those instances with greater harmfulness than usefulness.

\subsubsection{Hybrid methods}

{\it Decremental
Reduction Optimization Procedure 3} (DROP3)~\citep{wilson2000reduction} starts by executing ENN to remove noisy instances from the training set. Let $\mathcal{T}'$ be the output of ENN. The reduced training set $S$ is initialised to $\mathcal{T}'$, and its instances are traversed in decreasing order according to their distance to their nearest enemy. This order increases the likelihood of removing inner instances from $S$ and keeping border instances. Each instance $d$ is removed from $S$ if the number of instances in $\mathcal{T}'$ that are correctly classified with $S \setminus \{d\}$ is at least the same as the number of instances
in $\mathcal{T}'$ that are correctly classified with $S$.

{\it Iterative Case Filtering} (ICF)~\citep{brighton2002advances} first applies ENN to remove noisy instances.
Let $\mathcal{T}'$ be the output of ENN.
For each instance $d$ in $\mathcal{T}'$,
ICF computes its local set and compares its size, referred to as reachable$(d)$, with the number of instances $d'$ in $\mathcal{T}'$ such that $d$ belongs to the local set of $d'$, referred as coverage$(d)$. ICF removes from $\mathcal{T}'$ those instances $d$ such that reachable$(d) >$ coverage$(d)$.

{\it Local Set Border Selector} (LSBo)~\citep{leyva2015three} starts by executing LSSm to remove noisy instances.
Then, it computes the local sets over the reduced training set produced by LSSm. The reduced training set $S$ is initially empty. Instances in the reduced training set produced by LSSm are traversed in ascending order of the cardinality of their local sets. Each instance is added to $S$ if none of the instances in its local set is already in $S$. LSBo aims to keep the instances within the boundaries of each class. Its underlying idea is that instances in the boundaries usually have small local sets and, therefore, ordering them in ascending order of their local set sizes will usually keep them.

\section{Preliminaries}
\label{sec:preliminaries}

\subsection{Machine learning concepts}

The input of the CS/I selection problem is a training set composed of $n$ data instances (or samples), $\mathcal{T}= \{s_1, \ldots, s_n\}$. For each data instance $s_i$ we know a vector of $m$ features $f_i= (f_{i,1}, \ldots, f_{i,m}) \in F_1 \times \ldots \times F_m$ and a class $c_i \in \mathcal{C}$ to which the instance belongs. The set of possible classes $\mathcal{C}$ is assumed to be finite and known.

Given a (possibly reduced) training set $\mathcal{T}$, a classifier $K$ outputs a function $K(\mathcal{T}): F_1 \times \ldots \times F_m \rightarrow \mathcal{C}$ that estimates the class of a new instance given its vector of features. The {\it accuracy} of a classifier can be estimated with a test set $\mathcal{T}_s= \{s'_1, \ldots, s'_p\}$, which should be disjoint from the training set. For each data instance $s'_i$ in the test set, the corresponding vector of features $f'_i \in F_1 \times \ldots \times F_m$ and class $c'_i \in \mathcal{C}$ are also known. The accuracy of the classifier $K$ for training set $\mathcal{T}$ is then defined as $\frac{|\{s'_i \in \mathcal{T}_s: K(\mathcal{T})(f'_i)= c'_i\}|}{|\mathcal{T}_s|}$.

It is often the case that only a training set is available. In such cases, the accuracy of the classifier can be estimated with {\it $k$-fold cross-validation}. The data instances in the training set are randomly allocated in $k$ mutually disjoint subsets (the folds) $\mathcal{T}_1, \ldots, \mathcal{T}_k$ of equal size ($k=10$ is typical). Then, we evaluate the classifier in $k$ iterations. At each iteration, one fold plays the role of the test set while the other folds play the role of the training set. The accuracy of the classifier is then computed as $\frac{\sum_{j=1}^k|\{s_i \in \mathcal{T}_j: K(\mathcal{T} \setminus \mathcal{T}_j)(f_i)= c_i\}|}{|\mathcal{T}|}$.

We are also interested in the {\it reduction} achieved by CS/I selection algorithms, defined as {\it reduction}$= \frac{|\mathcal{T}|-|\mathcal{T}_r|}{|\mathcal{T}|}$, where $\mathcal{T}_r$ is the reduced dataset obtained with a certain algorithm. When we use $k$-fold cross-validation, we average the reduction values obtained from each cross-validation iteration.

Finally, we formalize the concept of {\it local set} that we presented in the Introduction and has been used before in the state of the art of instance selection~\citep{brighton2002advances,caises2011combining,leyva2013knowledge,leyva2015three}. We first introduce the concept of {\it nearest enemy}. Given a distance $d$, the
nearest enemy $ne(s_i)$
of each instance $s_i$
is the instance of a different class closest to $s_i$. The local set $\ell s$ of an instance $s_i$ is defined as the set of instances closer to $s_i$ than its nearest enemy.

\begin{equation}
\ell s(s_i)= \{s_j: d(s_i,s_j) < d(s_i,ne(s_i))\}
\end{equation}

It follows from the definition that all the instances in the local set of an instance $s_i$ belong to the same class as $s_i$.

\subsection{Social choice concepts}

An approval-based multi-winner election can be represented with a tuple $(N,C,\mathcal{A},t)$, where $N= \{v_1, \ldots, v_n\}$ is the set of $n$ {\it voters}, $C$ is the set of {\it candidates}, $\mathcal{A}= (A_1, \ldots, A_n)$ is the {\it ballot profile}, and $t$ is the {\it target committee size}. Approval voting means that voters do not order candidates according to their preferences, but just select the candidates they approve of. Thus, the ballot $A_i$ of a voter $v_i$ is a subset of the set of candidates $C$.

We depart from most work on approval-based multi-winner voting by allowing the actual committee size to be less than $t$. Thus, an approval-based multi-winner voting rule $R$ takes as input an election $(N, C,\mathcal{A},t)$ and outputs
a nonempty set of winners or committee $W \subseteq C$ of size less than or equal to $t$. We assume that voting rules are resolute. In our experiments, we break ties by selecting the data instance that comes first in the training set.

We now review two representation axioms that have been proposed in the literature: EJR, proposed by~\citet{aziz:scw}, and PJR, proposed by~\citet{pjr-aaai}.
Given a positive integer $\ell\leq t$, we say that a subset of the voters $N^*$ is {\em $\ell$-cohesive} if $|N^*| \geq \ell \frac{n}{t}$ and $|\bigcap_{v_i \in N^*} A_i| \geq \ell$.
The intuition behind $\ell$-cohesiveness is that each committee member must represent, roughly, $\frac{n}{t}$ voters. Thus, a group of voters of size at least $\ell \frac{n}{t}$ that approves of some $\ell$ candidates in common `deserves' to have $\ell$ representatives in the committee.

\begin{definition}
{\bf Extended/Proportional justified representation} \\
{\bf (EJR/PJR)}\footnote{In the original definitions, it was required that the size of the set of candidates $W$ is $t$.}
  Consider a ballot
  profile $\mathcal{A}= (A_1, \dots, A_n)$ over a candidate set $C$
  and a target committee size $t$.
  A set of candidates $W$ of size less than or equal to $t$
  is said to provide {\em $\ell$-extended justified representation}, $\ell$-EJR
  (respectively {\em $\ell$-proportional justified representation}, $\ell$-PJR),
  for $(N,C,\mathcal{A}, t)$ if there does not exist an $\ell$-cohesive set of voters
  $N^*$ such that $|A_i \cap W| < \ell$ for every $v_i \in N^*$ (respectively, such that $|W \cap \cup_{v_i \in N^*} A_i| < \ell$).
  We say that $W$ provides {\em extended justified
  representation (EJR)} for $(N,C,\mathcal{A}, t)$ if it provides $\ell$-EJR
  for $(N,C,\mathcal{A}, t)$ for every positive integer $\ell \leq t$; it provides {\em proportional justified
  representation (PJR)} for $(N,C,\mathcal{A}, t)$ if it provides $\ell$-PJR
  for $(N,C,\mathcal{A}, t)$ for every positive integer $\ell \leq t$.
\end{definition}

Intuitively, EJR requires that, for each $\ell$-cohesive group of voters, at least one group member approves of at least $\ell$ candidates in the committee. In contrast, PJR requires that, for each $\ell$-cohesive group of voters, at least $\ell$ candidates in the committee are approved of by some group members (although it is not required that the same member of the group approves of all the $\ell$ candidates).

We say that a voting rule satisfies an axiom if it always outputs committees that provide such an axiom.

We now introduce the voting rules that we have used in our experiments: {\it Simple EJR} (SEJR), the method of {\it Equal Shares} (ES), and {\it seq-Phragm\'en} (SeqP). SEJR was initially proposed by~\citet{SEL17a} and later, independently rediscovered and renamed as the Greedy Justified Candidate Rule by~\citet{brill2023robust}. It is an extreme case of the {\it EJR Exact} family of voting rules~\citep{aziz2018complexity,SANCHEZFERNANDEZ2025106576}. ES has been proposed by~\citet{peters2020proportionality}, and is also an instance of the EJR-Exact family~\citep{SANCHEZFERNANDEZ2025106576}. SeqP~\citep{phragmen:p1} was proposed by the Swedish mathematician Lars Edvard Phragm\'en at the end of the 19\textsuperscript{th} century\footnote{See~\citep{2016arXiv161108826J} for an overview on Phragm\'en rules.}. All the voting rules that belong to the EJR Exact family satisfy EJR~\citep{aziz2018complexity}. SeqP satisfies PJR but it fails EJR~\citep{brill2024phragmen}.
All these rules are polynomial-time computable and iterative
algorithms that start with an empty set of winners $W$ and add a new candidate to $W$ at each iteration, until a stop condition is met.

We review here the operation of SEJR. A detailed description of ES and SeqP is deferred to the appendix. SEJR uses the notion of {\it EJR+ demand}.

\begin{definition}[\citet{SANCHEZFERNANDEZ2025106576}]
\label{def:ejr+demand}
    The EJR+ demand $\textrm{d}(c,W)$ of a candidate $c \in C \setminus W$ with respect to a set of candidates $W \subseteq C$ such that $|W| \leq t$,  is the maximum non-negative integer such that a group of voters $N'$ exists satisfying that (i) $c$ belongs to $A_i$ for each voter $i$ in $N'$; (ii) $|N'| \frac{t}{n} \geq \textrm{d}(c,W)$; and (iii) $|A_i \cap W| < \textrm{d}(c,W)$ for each voter $i$ in $N'$. If conditions (i), (ii), and (iii) are not satisfied for any $N'$ and any positive value of $\textrm{d}(c,W)$, then $\textrm{d}(c,W)= 0$.
\end{definition}

At each iteration,
SEJR selects the candidate $c$ with maximum EJR+ demand,
until the EJR+ demand of all the remaining candidates is $0$.
Pseudocode for computing SEJR is given in~\citep[p.~25]{SEL17a}.

An interesting property of SEJR and ES is that they provide {\it incomplete committees}: given an approval-based multi-winner election $(N, C,\mathcal{A},t)$, they usually output committees of size less than $t$. In the first experiment, we will need the committee's size to be exactly $t$. We achieve this by following the approach proposed by~\citet{peters2020proportionality} for ES: we first run SEJR or ES, and then complete the committee by running the rule SeqP to add additional candidates to the set of winners.

On the other hand, in the second experiment, we take advantage of this feature of SEJR and ES to compute the winning committee for elections in which the target committee size $t$ is even greater than the number of candidates that participate in the election, even though it is not possible for the actual committee to be larger than the set of candidates. The reason is that the value of $t$ also influences the guarantees offered by the EJR and PJR axioms. For instance, setting $t= 2n$ in our CS/I selection scenario,
where $n$ is the number of both voters and candidates, guarantees that each voter will have at least two of its preferred candidates in the winning committee produced by SEJR, and ES, as long as such voter approves of at least two candidates. In the rest of this paper, we will refer to rules that allow for arbitrarily large integer values of $t$ as {\it unconstrained target committee size} (UTCS) rules. The rule SeqP always ouputs a committee of size $t$, and therefore, it is not an UTCS rule.

SEJR and ES can potentially output empty committees if no candidate is approved of by at least $\lceil \frac{n}{t} \rceil$ voters. In such case, we would have assumed that they output a committee composed of only the most approved of candidate, although this has never happened in our experiments.

The worst-case time complexity of SEJR, ES, and SeqP is similar. It can be bounded by $\mathcal{O}(nmt)$ for SEJR and SeqP, and by $\mathcal{O}(n(m+\log n)t)$ for ES, where $n$ is the number of voters, $m$ is the number of candidates and $t$ is the target committee size.\footnote{A worst-case time complexity bound for SEJR has been obtained by~\citet{SEL17a}. Worst-case time complexity bounds for ES and SeqP are obtained in the appendix.}
We also note that all these rules compute at each iteration a certain value or score for each candidate $c$ (for instance, in the case of SEJR, such score is the EJR+ demand of each candidate),
and select the candidate with better score. The score of each candidate can be computed in parallel for all the rules, so if we have $m$ processing units available, we can compute the outcome of the three rules in $\mathcal{O}(nt)$. In the appendix we also discuss the memory complexity of these rules.

\section{A social choice model for CS/I selection\label{sec:model}}

The notion of local set, introduced in the previous section, allows us to identify, for each instance $s_i$, a set of instances  $\ell s(s_i)$ of the same class in a hypersphere where no instance of another class is present. Our idea is to consider the instances in $\ell s(s_i)$ as possible representatives of $s_i$. In other words, the instances that $s_i$ will approve of must belong to its local set. When running the election, the vote of $s_i$ can be seen as a demand that the hypersphere where the instances in $\ell s(s_i)$ are placed is well represented in the reduced dataset and therefore assigned to the class to which $s_i$ belongs. If several other instances have local sets that intersect with the hypersphere of $s_i$, then $s_i$, together with those other instances, will receive a number of representatives according to their size that will lay in the area of interest of those instances. We note that a similar idea is expressed by~\citep{brighton2002advances}. In the terminology employed by~\citep{brighton2002advances}, an instance is `adaptable' by the instances that belong to its local set.

Instances near the class boundary will have small local sets that will correspond to a small hypersphere. To have such instances well represented, it is necessary that instances at the border of each class are present in the reduced dataset, which in turn allows us to identify the area corresponding to each class in the dataset. Inner instances will have large local sets that may include some instances in the border, so they can be happy to be represented by them. In summary, we expect that the reduced dataset obtained by running a representative voting rule over an approval-based multi-winner election in which each instance approves of (a subset of) the instances in its local set will identify well the borders of each class, which in turn will allow us to obtain good accuracy with classifiers based on identifying instances in the border of each class (the support vectors) like SVMs.
However, our approach is not limited to identifying the borders of each class, because the proportionality criteria we use ensures that the different areas of each class are represented according to the number of instances with such areas in their local sets.
These ideas will be illustrated graphically in the next section.
Finally, to add an additional degree of flexibility, we will initially allow each instance to approve of a subset of its local set (and not necessarily its entire local set).

Definition~\ref{def:model}
formalises
our social choice model for CS/I selection,
which defines a family of approval-based multi-winner elections derived from a given training set:

\begin{definition}
\label{def:model}
Consider a training set $\mathcal{T}= \{s_1, \ldots, s_n\}$.
We say that an approval-based multi-winner election $(N,C,\mathcal{A},t)$ is local-set-derived from $\mathcal{T}$ if (i) $N=C=\mathcal{T}$, and (ii) the approval ballot $A_i$ of each instance $s_i$ is a subset of the local set of $s_i$: $A_i \subseteq \ell s(s_i)$.
\end{definition}

In Sections~\ref{sec:graphic},~\ref{sec:core-set}, and~\ref{sec:carbonera}, we define the approval ballot $A_i$ of each instance $s_i$ as all the instances in its local set except itself, $A_i = \ell s(s_i) \setminus \{s_i\}$. We have selected this value inspired by the design of LSSm~\citep{leyva2015three}. As discussed in~\citep{leyva2015three}, instances that do not belong to the local set of any other instance may be noisy and induce mistakes in the classifier. Our choice for $A_i$ guarantees that they are not selected by our voting rules.

Given an approval-based multi-winner election $(\mathcal{T}, \mathcal{T}, \mathcal{A}, t)$, local-set-derived from a training set $\mathcal{T}$, we can compute a reduced training set $\mathcal{T}_r$
with voting rule $R$ as the set of winners that $R$ outputs for election $(\mathcal{T}, \mathcal{T}, \mathcal{A}, t)$.

\begin{equation}
\mathcal{T}_r= R(\mathcal{T}, \mathcal{T}, \mathcal{A}, t)
\end{equation}

Our first result for local-set-derived elections is a theoretical result for KNN classifiers. First, we recall the operation of a KNN classifier. Given the vector of features of a new instance whose class is unknown, a KNN classifier computes the distance between such a vector and the vectors of features of the instances in the training set. The class of the new instance is then estimated by using a majority rule: the most frequent class of the K instances in the training set closer to the new instance is selected. The value of K is usually odd to avoid ties, with typical values of K$=1$ or K$=3$.

\begin{theorem}
\label{th:pjr-knn}
    Fix an odd positive integer K. For any training set $\mathcal{T}= \{s_1, \ldots, s_n\}$, consider an approval-based multi-winner election $(\mathcal{T}, \mathcal{T}, \mathcal{A},t)$, with $t= \frac{\textrm{K}+1}{2} |\mathcal{T}|$, local-set-derived from $\mathcal{T}$ and a UTCS rule $R$ that satisfies $\frac{\textrm{K}+1}{2}$-PJR. Then, a KNN classifier trained with the reduced training set $\mathcal{T}_r$ obtained as the output of rule $R$ for approval-based multi-winner election $(\mathcal{T}, \mathcal{T}, \mathcal{A},t)$ will classify correctly any instance $s_i$ of the original training set $\mathcal{T}$ such that the approval set $A_i$ of $s_i$ has a size of at least $\frac{\textrm{K}+1}{2}$.
\end{theorem}

\begin{proof}
    We first note that, under the conditions established in this theorem, each group of voters $N^*$ consisting of a single voter $s_i$ that approves of at least $\frac{\textrm{K}+1}{2}$ candidates is a $\frac{\textrm{K}+1}{2}$-cohesive group of voters, because $|N^*|= |\{s_i\}|= 1= \frac{\textrm{K}+1}{2} \frac{|\mathcal{T}|}{t}$, and $|A_i| \geq \frac{\textrm{K}+1}{2}$.

    Since the rule $R$ satisfies $\frac{\textrm{K}+1}{2}$-PJR, this implies that
    $|\mathcal{T}_r \cap \cup_{s_j \in N^*} A_j|= |\mathcal{T}_r \cap A_i| \geq \frac{\textrm{K}+1}{2}$
    for the committee $\mathcal{T}_r$ that $R$ outputs for election $(\mathcal{T}, \mathcal{T}, \mathcal{A},t)$. Now, since all the instances in $A_i$ belong to the local set of $s_i$, there are at least $\frac{\textrm{K}+1}{2}$ instances in $\mathcal{T}_r$ that are closer to $s_i$ than the nearest enemy of $s_i$ in $\mathcal{T}$. However, $\mathcal{T}_r$ is a subset of $\mathcal{T}$, and therefore, the nearest enemy of $s_i$ in $\mathcal{T}_r$ cannot be closer to $s_i$ than the nearest enemy of $s_i$ in $\mathcal{T}$. Therefore, there are at least $\frac{\textrm{K}+1}{2}$ instances in $\mathcal{T}_r$ of the same class as $s_i$ among its K nearest neighbours, and thus KNN would classify $s_i$ correctly with the reduced training set $\mathcal{T}_r$ output by rule $R$.
\end{proof}

\section{Graphical representation}
\label{sec:graphic}

It is instructive to examine the effect of applying voting rules to a dataset graphically.
In order to do that,
we apply Principal Component Analysis (PCA)
to reduce the dimensions of the feature space to the two directions that capture the most significant variation in the data values.

\begin{figure*}[thb]
    \centering
    \begin{subfigure}{0.32\textwidth}
        \includegraphics[width=\textwidth]{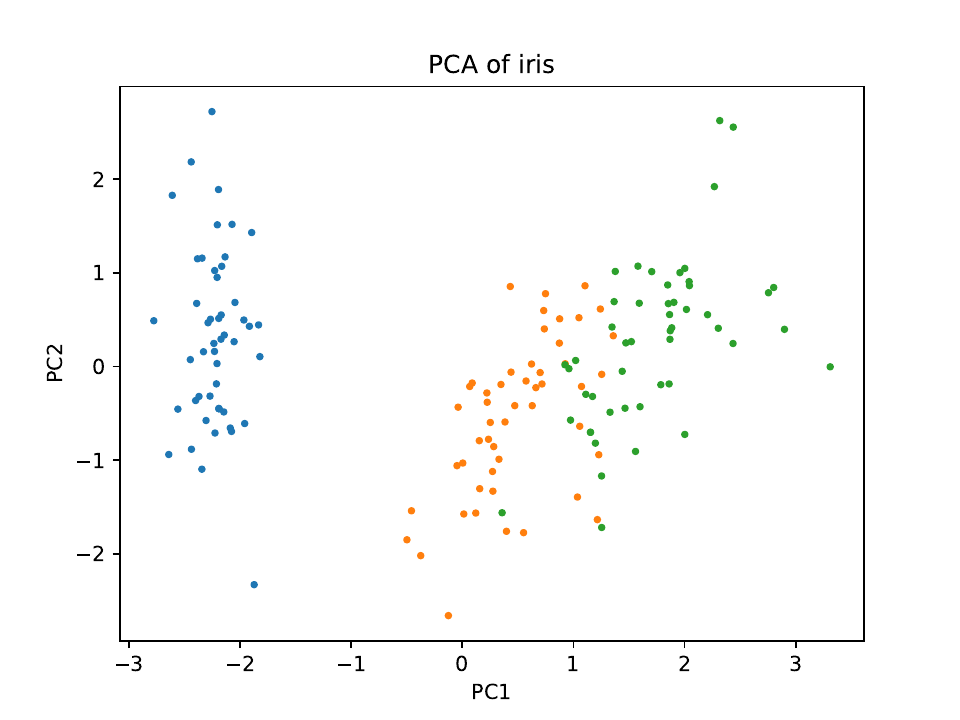}
        \caption{Original Iris dataset}
        \label{fig:pca-iris-ori}
    \end{subfigure}
    \hfill
    \begin{subfigure}{0.32\textwidth}
        \includegraphics[width=\textwidth]{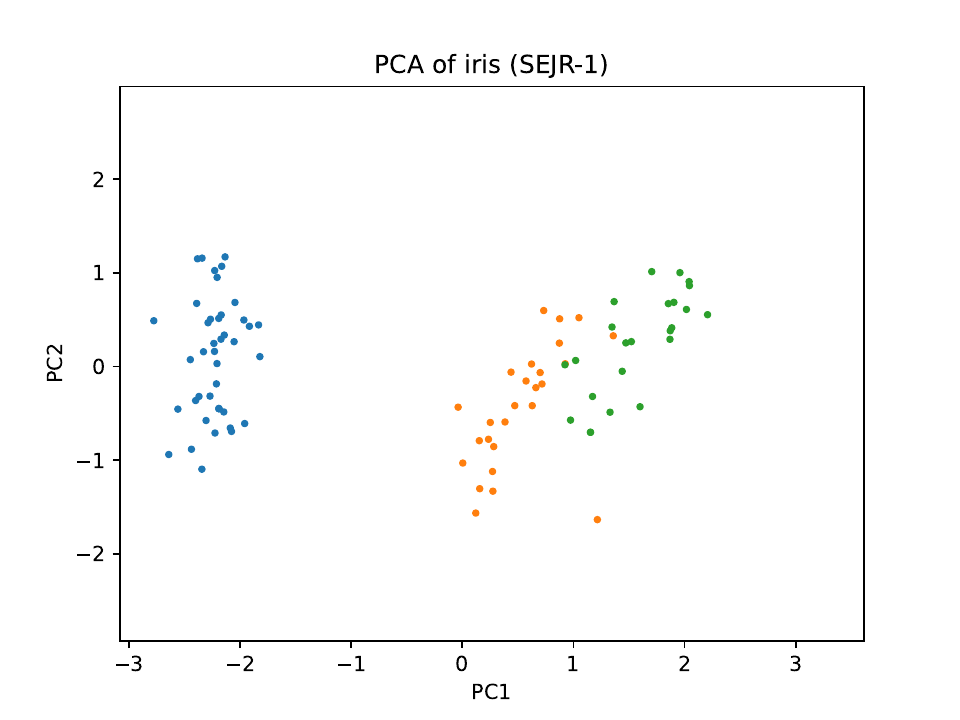}
        \caption{Reduced with SEJR-$1$}
        \label{fig:pca-iris-sejr-1}
    \end{subfigure}
    \hfill
    \begin{subfigure}{0.32\textwidth}
        \includegraphics[width=\textwidth]{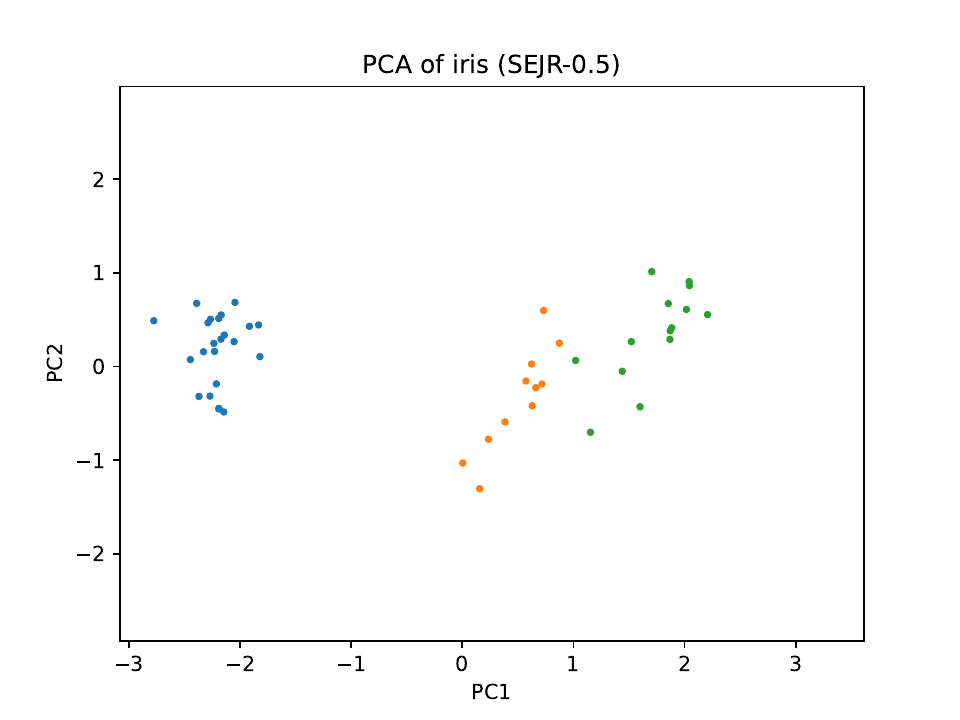}
        \caption{Reduced with SEJR-$0.5$}
        \label{fig:pca-iris-sejr-0.5}
    \end{subfigure}
    \caption{PCA plots of the Iris dataset}
    \label{fig:pca-iris}
\end{figure*}

\begin{figure}[thb]
    \centering
    \begin{subfigure}{0.32\columnwidth}
        \includegraphics[width=\textwidth]{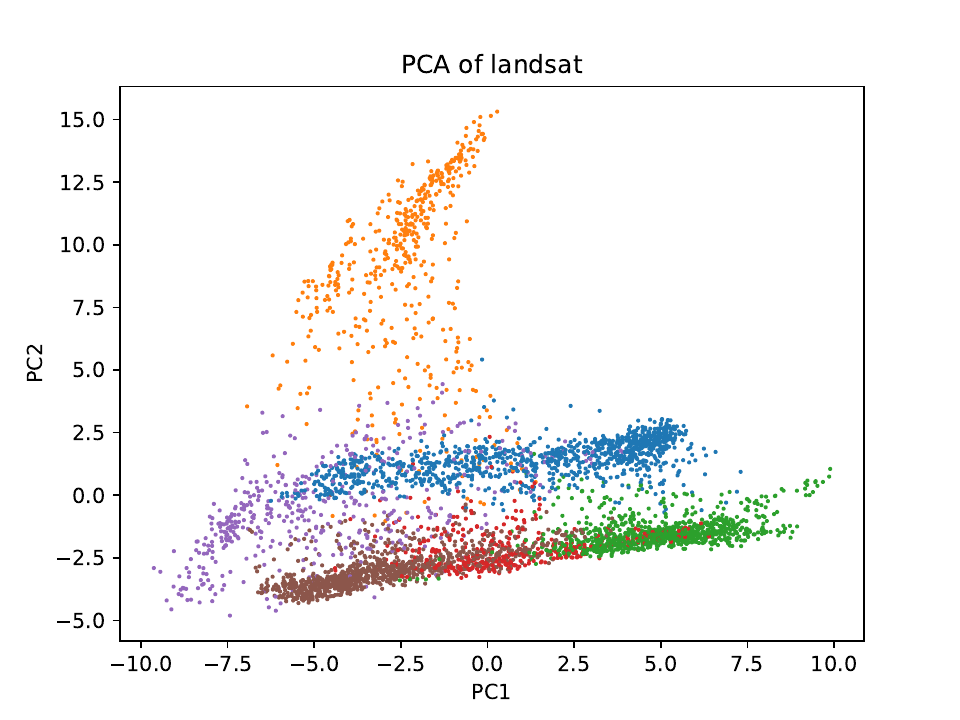}
        \caption{Original dataset}
        \label{fig:pca-landsat-ori}
    \end{subfigure}
    \hfill
    \begin{subfigure}{0.32\columnwidth}
        \includegraphics[width=\textwidth]{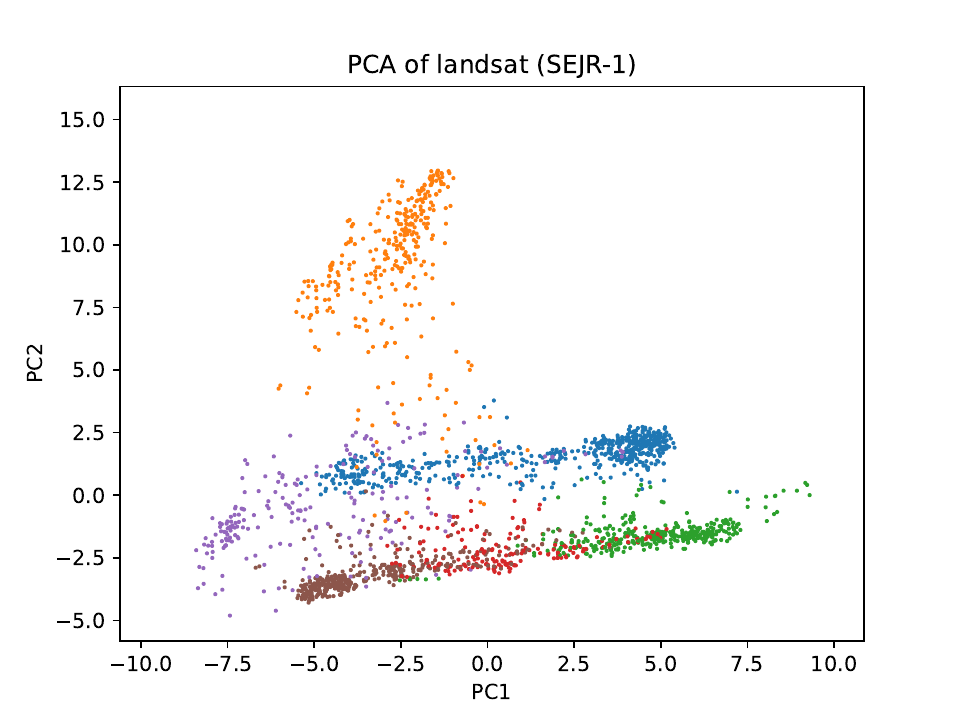}
        \caption{Reduced with SEJR-$1$}
           \label{fig:pca-landsat-sejr-1}
    \end{subfigure}
    \hfill
    \begin{subfigure}{0.32\columnwidth}
        \includegraphics[width=\textwidth]{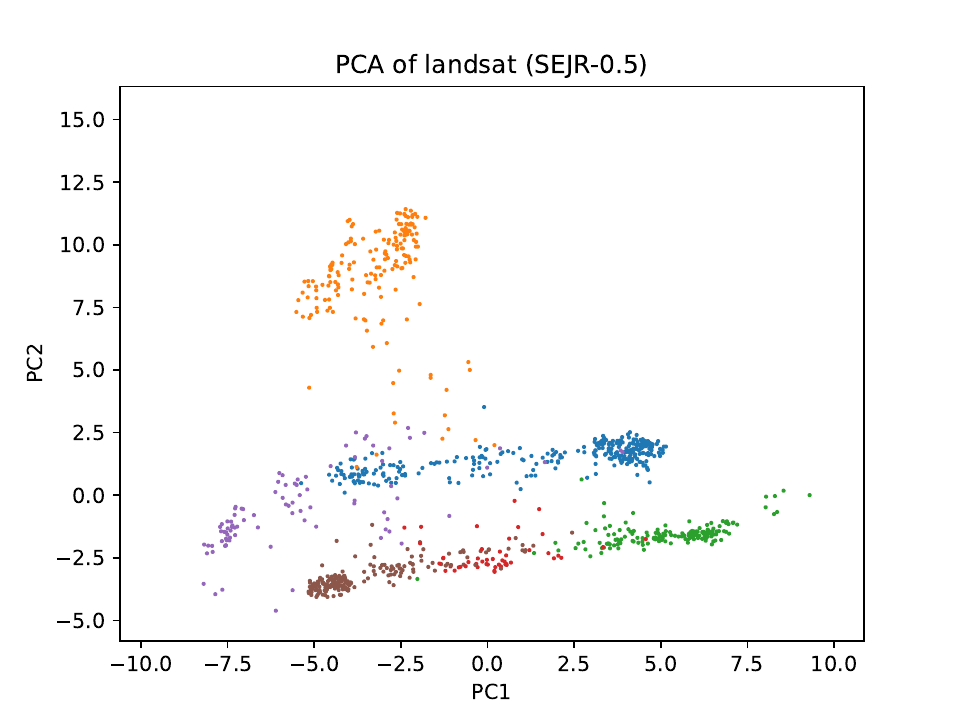}
        \caption{Reduced with SEJR-$0.5$}
        \label{fig:pca-landsat-sejr-0.5}
    \end{subfigure}
    \caption{PCA plots of the Landsat dataset}
    \label{fig:pca-landsat}
\end{figure}

Figures~\ref{fig:pca-iris} and~\ref{fig:pca-landsat} show several plots obtained with two datasets from the UCI Machine Learning Repository~\citep{Dua:2019}: Iris, a classic small dataset of $150$ instances that belong to three classes, and Landsat, consisting of $3\times 3$ pixels from $4,435$ satellite images belonging to six classes. All the features of both datasets are numerical.
In both figures,
and in the rest of this paper,
we use labels that indicate the voting rule used and the value of $\frac{t}{n}$. For instance, SEJR-$1$ means rule SEJR and $\frac{t}{n}=1$.

We discuss the effect of applying representative voting rules to a dataset considering three types of instances: (i) instances placed in an area where instances of several classes are intermixed; (ii) instances close to the border of a class, but such that there is no instance of a different class between the instance and the interior of the area occupied by the class; and (iii) instances that are far away from the border of the class.

Type (iii) instances are good candidates to be removed because they are not approved of by those instances of the same class closer to the class border. In contrast, type (ii) instances are primarily kept in the dataset because all the inner instances approve of them.
Type (i) instances may have low approval scores because inner instances can have their nearest enemy closer than them. However, the behaviour of the rules for these instances is different from type (iii) instances because the latter usually approve of the instances of type (ii) and, therefore, they may be satisfied with being represented by them. However, instances in intermixed areas may not approve of type (ii) instances due to the reduced area of their local sets. Thus, they may demand to be represented by
other type (i) instances in the reduced training set.

Figure~\ref{fig:pca-iris} shows plots for the original Iris dataset reduced with SEJR-$1$ and SEJR-$0.5$. Comparing the plots for the original dataset and SEJR-$1$, we observe that the instances of the blue and green classes that are further away from the orange class get removed. In the case of the orange class, its gap to the blue class makes orange instances closer to the blue class have large local sets and, thus, the most approved of orange instances are type (ii) instances close to the border with the green class. This can be seen as if the orange class were worried about `defending' its border with the green class, while being less worried about the already clean border with the blue class. The effect of removing each instance from its approval ballot is also apparent: all the instances surrounded by other class instances are removed.

At a coarser level, due to the higher number of instances, a similar behaviour can be observed for the Landsat dataset in Figure~\ref{fig:pca-landsat}. For example, we can observe that the orange class's tail is reduced with SEJR-$1$ and even more with SEJR-$0.5$.

\section{Experiment I: experiment with neural networks}
\label{sec:core-set}

Our first experiment
consisted of evaluating the performance of the proposed reduction algorithms
on the CIFAR-10 and CIFAR-100~\cite{krizhevsky2009learning} datasets,
which are composed of 60,000
(50,000 for training and 10,000 for testing)
small 32x32 colour images each,
divided into 10 and 100, respectively, balanced classes.

We compared the SEJR, ES and seqP rules
with the herding~\cite{chen2010super}, k-Means and k-Center core set algorithms,
and the random selection baseline.
The feature vectors needed by those algorithms
were computed by training
a ResNet50 neural network,
initialised with random weights,
for five epochs
with all the samples in the training set,
separately for each dataset.
We extracted 2,048 features
for each sample.
Algorithms needing sample-to-sample distance computation used Euclidean distances.

Following the same approach as in~\cite{cui2022dc},
each algorithm was configured
to produce balanced fixed-size reductions
of the training set
with two different sizes:
50 images per class
and 10 images per class.
That means that,
out of 50,000 samples,
a total of 5,000 and 1,000 images were selected for CIFAR-100,
whereas 500 and 100 images were selected for CIFAR-10.

Also similarly to~\cite{cui2022dc},
different neural network architectures
(ConvNet, MLP, ResNet18 and ViT)
were trained for 300 epochs,
starting from random weights,
with the samples selected from the training set
by the different algorithms.
These networks were set-up with the same configuration as in~\cite{cui2022dc}.
Because of the random initialisation of the networks,
training was repeated 40 times for each reduction,
except for the $k$-Center and random algorithms,
for which the selection itself is random.
For these two approaches,
8 selections were produced
and network training was repeated 5 times with each one.

The accuracy of each network,
measured on the test set
as described in Section~\ref{sec:preliminaries},
was computed,
obtaining 40 values for each combination of
dataset, algorithm and network.
They were averaged,
and their 95\% confidence intervals were computed.
Table~\ref{tab:exp-1-accuracy} shows them.

\begin{table}[t]
    \scriptsize
    \setlength{\tabcolsep}{3pt}     
    \renewcommand{\arraystretch}{0.95}

  \begin{tabular}{M{0.1cm}rlrrrrrrrr}
    & \multicolumn{1}{c}{IPC} & \multicolumn{1}{c}{Classifier} & \multicolumn{1}{c}{SEJR} & \multicolumn{1}{c}{ES} & \multicolumn{1}{c}{SeqP} & \multicolumn{1}{c}{Herding} & \multicolumn{1}{c}{k-Means} & \multicolumn{1}{c}{k-Center} & \multicolumn{1}{c}{Random} \\
    \hline
    \multirow{8}{*}{\makecell{\rotatebox{90}{CIFAR-100}}} & \multirow[t]{4}{*}{50} & ConvNet & $33.57 \pm 0.09$ & $33.58 \pm 0.11$ & $\textbf{33.62} \pm 0.09$ & $32.74 \pm 0.11$ & $31.10 \pm 0.11$ & $23.83 \pm 0.11$ & $30.08 \pm 0.13$ \\

    &  & MLP & $\textbf{17.90} \pm 0.08$ & $17.86 \pm 0.08$ & $17.89 \pm 0.06$ & $15.80 \pm 0.06$ & $14.67 \pm 0.08$ & $9.83 \pm 0.10$ & $14.53 \pm 0.10$ \\

    & & ResNet18 & $\textbf{19.32} \pm 0.17$ & $19.02 \pm 0.13$ & $19.21 \pm 0.15$ & $14.48 \pm 0.19$ & $14.95 \pm 0.16$ & $10.68 \pm 0.12$ & $14.17 \pm 0.28$ \\

    & & ViT & $\textbf{21.77} \pm 0.11$ & $21.64 \pm 0.10$ & $21.48 \pm 0.12$ & $18.15 \pm 0.11$ & $17.49 \pm 0.11$ & $11.45 \pm 0.13$ & $16.60 \pm 0.14$ \\
    \cline{2-10}
     & \multirow[t]{4}{*}{10} & ConvNet & $19.52 \pm 0.08$ & $\textbf{19.56} \pm 0.08$ & $ 19.55 \pm 0.08$ & $17.28 \pm 0.09$ & $15.37 \pm 0.08$ & $8.98 \pm 0.12$ & $14.44 \pm 0.14$ \\

    &  & MLP & $\textbf{12.65} \pm 0.06$ & $12.61 \pm 0.06$ & $12.60 \pm 0.06$ & $11.17 \pm 0.06$ & $8.92 \pm 0.06$ & $5.05 \pm 0.05 $ & $8.99 \pm 0.10$ \\

    & & ResNet18 & $8.42 \pm 0.12$ & $8.56 \pm 0.11$ & $\textbf{8.61} \pm 0.10$ & $5.55 \pm 0.11$ & $5.50 \pm 0.10$ & $3.14 \pm 0.08$ & $5.08 \pm 0.11$ \\

            & & ViT & $9.98 \pm 0.28$ & $9.78 \pm 0.33$ & $ \textbf{10.05} \pm 0.25$ & $6.83 \pm 0.11$ & $6.85 \pm 0.20$ & $3.89 \pm 0.11$ & $5.97 \pm 0.14$ \\

    \hline

    \multirow{8}{*}{\makecell{\rotatebox{90}{CIFAR-10}}} & \multirow[t]{4}{*}{50} & ConvNet & $46.56 \pm 0.12$ & $46.32 \pm 0.12$ & $\textbf{46.63} \pm 0.11$ & $46.13 \pm 0.14$ & $44.61 \pm 0.09$ & $38.25 \pm 0.3$ & $43.72 \pm 0.22$ \\

    &  & MLP & $32.90 \pm 0.09$ & $32.65 \pm 0.09$ & $\textbf{32.93} \pm 0.07$ & $30.72 \pm 0.10$ & $28.87 \pm 0.08$ & $25.71 \pm 0.26$ & $28.87 \pm 0.08$ \\

    & & ResNet18 & $\textbf{32.49} \pm 0.26$ & $32.31 \pm 0.24$ & $32.43 \pm 0.23$ & $28.56 \pm 0.23$ & $28.69 \pm 0.24$ & $23.86 \pm 0.39$ & $27.98 \pm 0.23$ \\

            & & ViT & $31.56 \pm 0.23$ & $31.77 \pm 0.22$ & $\textbf{32.12} \pm 0.29$ & $25.35 \pm 0.26$ & $25.15 \pm 0.29$ & $21.37 \pm 0.24$ & $25.29 \pm 0.30$ \\
    \cline{2-10}
     & \multirow[t]{4}{*}{10} & ConvNet & $32.72 \pm 0.16$ & $32.89 \pm 0.13$ & $\textbf{32.92} \pm 0.11$ & $26.69 \pm 0.19$ & $31.15 \pm 0.20$ & $25.14 \pm 0.35$ & $26.81 \pm 0.63$ \\

    &  & MLP & $25.74 \pm 0.09$ & $\textbf{26.62} \pm 0.10$ & $26.54 \pm 0.10$ & $22.85 \pm 0.12$ & $25.43 \pm 0.13$ & $20.96 \pm 0.43$ & $22.47 \pm 0.47$ \\

    & & ResNet18 & $25.07 \pm 0.22$ & $25.68 \pm 0.23$ & $\textbf{25.87} \pm 0.20 $ & $20.62 \pm 0.20$ & $23.14 \pm 0.27$ & $18.10 \pm 0.37$ & $20.08 \pm 0.41$ \\

            & & ViT & $ \textbf{23.10} \pm 0.11$ & $23.07 \pm 0.12$ & $22.98 \pm 0.11$ & $21.55 \pm 0.12$ & $21.54 \pm 0.14$ & $18.96 \pm 0.22$ & $20.02 \pm 0.50$ \\
\hline
  \end{tabular}
  \caption{Average accuracy results for experiment I,
    with 95\% confidence intervals.}
  \label{tab:exp-1-accuracy}
\end{table}

The SEJR, ES and SeqP rules
consistently obtain a better accuracy than the rest of the core set algorithms
across all the configurations,
except in one case, in which herding performs similarly to ES
but worse than the other two rules.
This is remarkable,
since other experiments in the state of the art~\cite{cazenavette2022dataset, zhao2023dataset,zhao2021dataset}
show that herding consistently performs better than other core set algorithms
they compare it to.

As we explained in the Introduction, SEJR and ES usually output incomplete committees, which we complete later using SeqP. We have also explored the possibility of optimising (increasing) the target committee size when running SEJR or ES to minimise the number of instances that should be added using SeqP, but this comes at the price of executing several SEJR or ES runs to find the optimal target committee size, thereby significantly increasing computation time.
The results obtained are similar to those reported here for a single run of SEJR or ES, except in the case of CIFAR-10 with 10 instances per class, in which the version with an optimized target committee size significantly improves accuracy. The results can be found in the appendix.

\begin{figure*}[thb]
    \centering
    \begin{subfigure}{0.49\textwidth}
        \includegraphics[width=\textwidth]{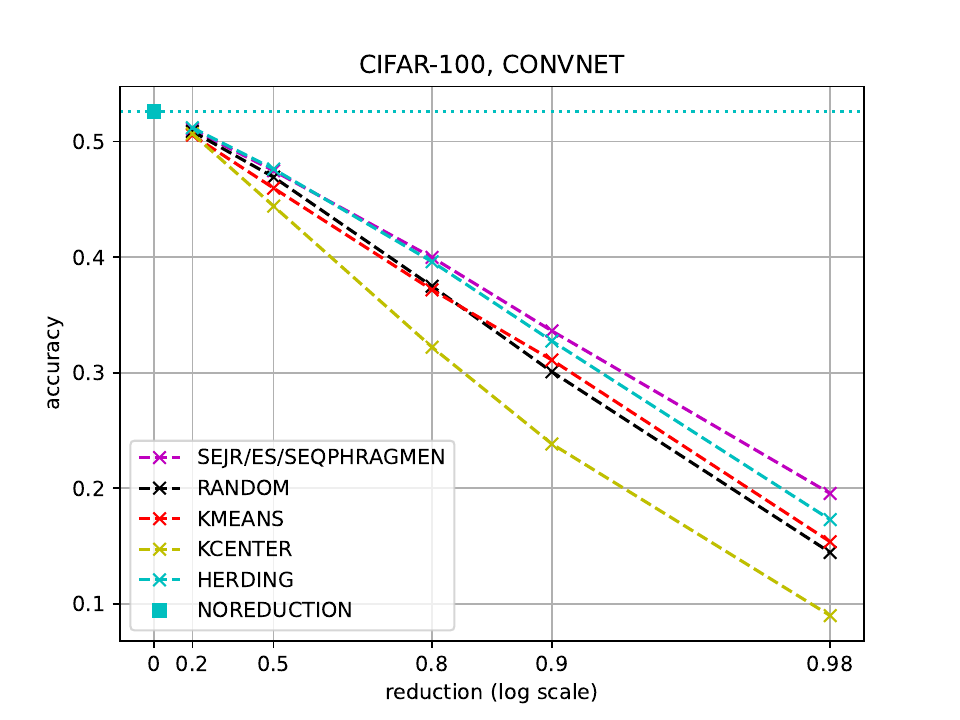}
    \end{subfigure}
    \hfill
    \begin{subfigure}{0.49\textwidth}
        \includegraphics[width=\textwidth]{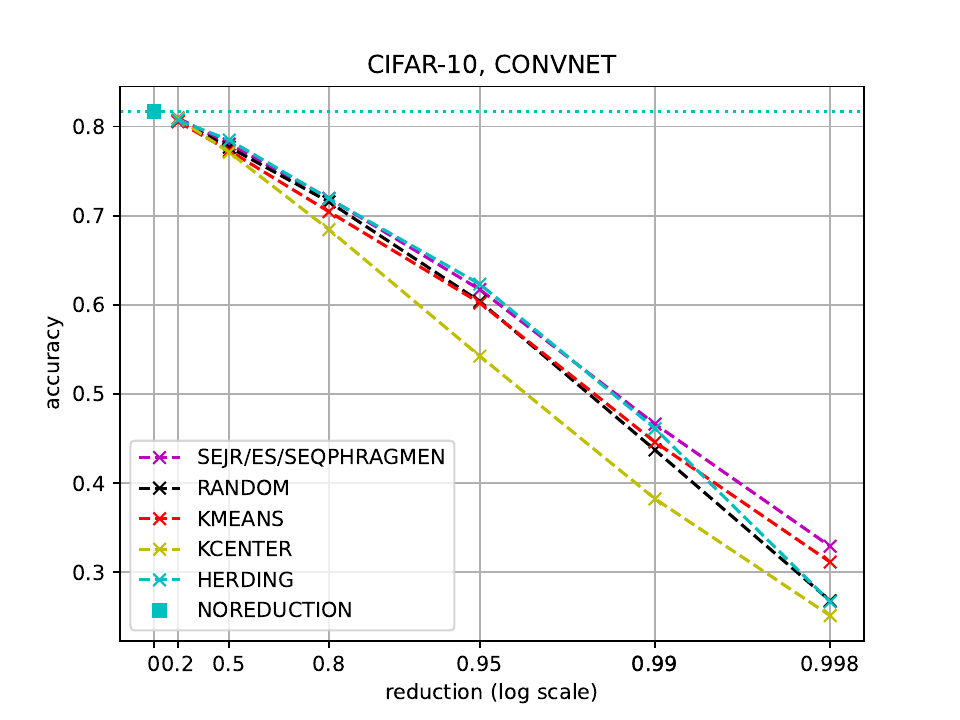}
    \end{subfigure}
    \caption{Accuracy vs. reduction curves
      with the ConvNet classifier,
      for CIFAR 100
      (trained with 400, 250, 100, 50 and 10 images per class)
      and CIFAR 10
      (trained with 4000, 2500, 1000, 250, 50 and 10 images per class),
      compared to training with the whole train split of the datasets
      (NOREDUCTION).
      For clarity,
      SEJR, ES and SeqP are shown as a single curve,
      since their accuracy values overlap
      at all the tested reduction ratios.
    }
    \label{fig:acc-red-cifar}
\end{figure*}

Figure~\ref{fig:acc-red-cifar} shows how accuracy drops
with growing reduction ratios.
Similarly to what is reported in~\cite{cui2022dc},
differences in accuracy between algorithms
are almost indistinguishable for small reduction ratios,
and more noticeable as these ratios increase.
In particular,
SEJR, ES and SeqP
produce bigger accuracy gains versus random selection
and the core set baselines
as reduction increases.

\begin{figure*}[thb]
  \centering
  \begin{subfigure}{0.49\textwidth}
    \includegraphics[width=\textwidth]{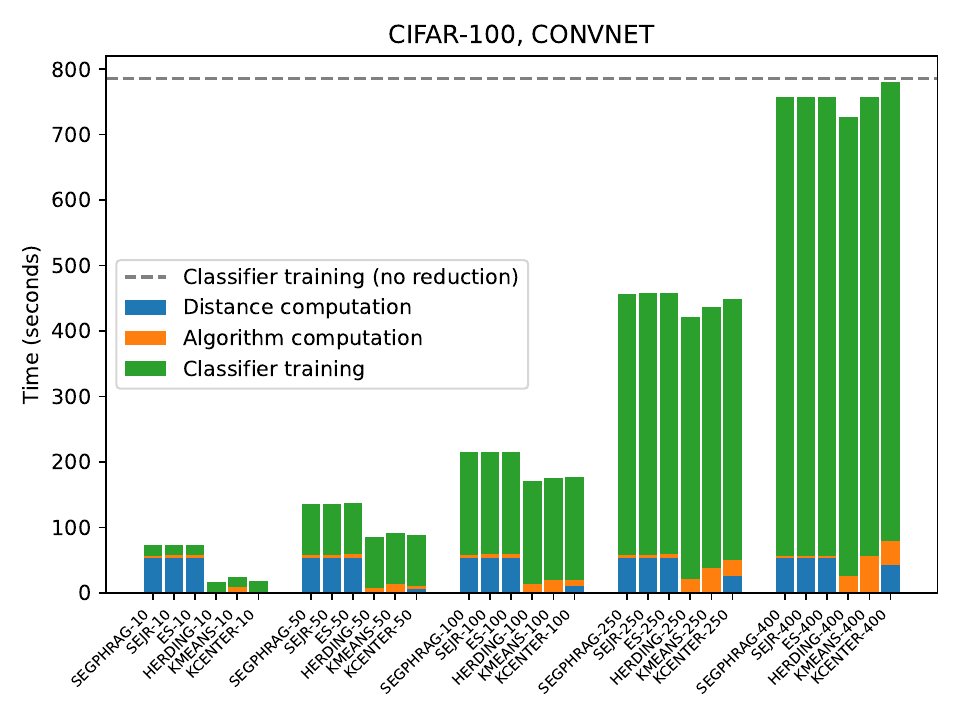}
  \end{subfigure}
  \hfill
  \begin{subfigure}{0.49\textwidth}
    \includegraphics[width=\textwidth]{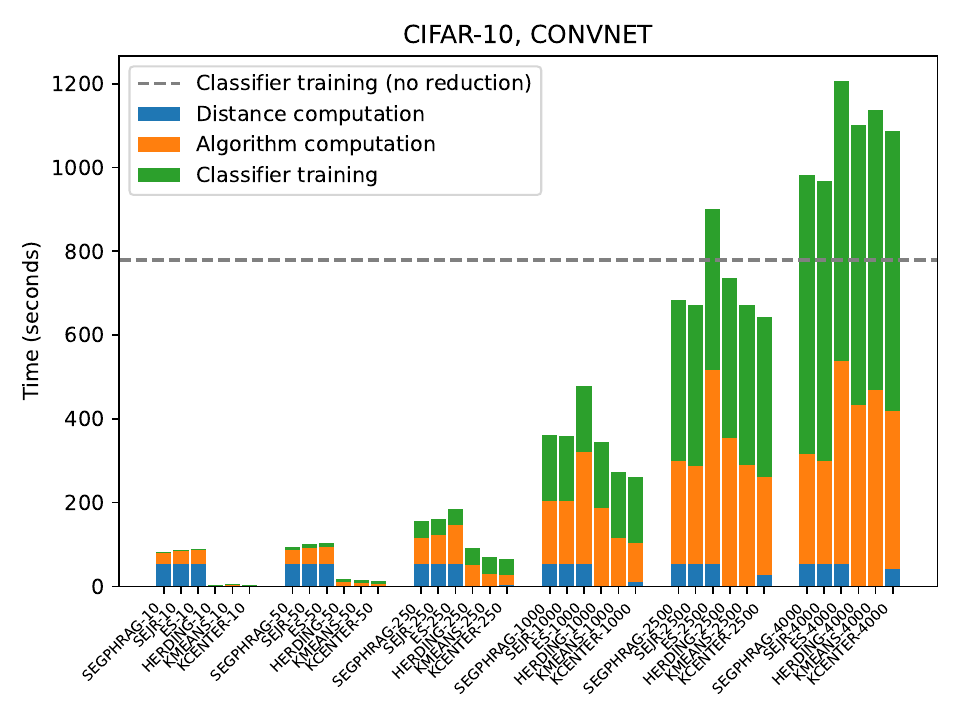}
  \end{subfigure}
  \caption{Total time needed to reduce the dataset and train
    the classifier for different values of images per class
    (10, 50, 100, 250 and 400 for CIFAR 100;
    10, 50, 250, 1000, 2500 and 4000 for CIFAR 10).}
  \label{fig:times-cifar}
\end{figure*}

Figure~\ref{fig:times-cifar} shows the amount of time
needed to run each reduction algorithm\footnote{
  Each data point is the average of 10 runs
  on a server with an AMD Ryzen 9 7900X 12-Core Processor,
  NVIDIA GeForce RTX 4090 GPU and 128 GB of RAM.
  Classifier training was run on the GPU,
  as well as distances computation.
  For k-means reduction, its multi-threaded implementation
  from the scikit-learn 1.7.2 library was used.
  The rest of the selection algorithms were implemented
  with single-threaded pure python code,
  with some of their operations,
  such as local sets computation,
  using the NumPy 2.3.4 library.
  Their computation times could be further reduced
  with a multi-threaded implementation,
  since the reduction of each class
  is independent from the other classes.
  },
including distances computation
for the algorithms that need them\footnote{Distances
  can be computed once for each dataset and cached to memory or a file.},
and for training the ConvNet classifier\footnote{The ConvNet
  classifier was trained with the same set-up used for obtaining
  the accuracy values of Table~\ref{tab:exp-1-accuracy}:
  300 epochs, starting from random weights.}
with the reduced dataset.
The time needed to train the classifier with the original dataset,
without reduction,
is also shown as a reference.
Computation times of SEJR, ES and SeqP
are worse for higher reduction rates
with respect to other core set algorithms
due to the computation of the distances they need.
Nevertheless,
these are the reduction rates at which they yield
greater accuracy gains with respect to them.
Results also show
that reducing the dataset pays off in terms of computation time
for most reduction rates,
even for a single training.
Computation savings would be greater in settings
in which the classifier needs to be trained several times
for the same reduction,
such as for hyper-parameter adjustment or continuous learning.

\section{Experiment II: experiment with classic machine learning classifiers}
\label{sec:carbonera}

In this experiment, we follow the typical methodology used in previous work on instance selection.
We use $k$-fold cross-validation with $k= 10$.
One fold is used for testing; the others are the initial training set.
Then, we build an approval-based multi-winner election with the instances in
the initial training set and run an approval-based multi-winner voting rule to
compute the reduced training set. The reduced training set is then used to train
a classifier, and the accuracy of the classifier is computed on the testing fold. The process is repeated so that each fold plays the testing role once.
The final accuracy is computed as described in Section~\ref{sec:preliminaries}.

We compare the results obtained with the voting rules considered in this paper with those reported by~\citet{carbonera2018efficient} for seven prominent instance selection algorithms: DROP3, ENN, ICF, LSBo, LSSm, LDIS, and ISDSP. In our experiments, we use the same 15 datasets (see Table~\ref{t:datasets}), the same distance (Euclidean) and evaluate the accuracy of the reduced training sets with the same classifiers adjusted with the same parameters as in~\cite{carbonera2018efficient}.
All the datasets have been obtained from the UCI Machine Learning Repository~\cite{Dua:2019}. All the features used in all the datasets are numerical.

\begin{table}[thb]
\begin{center}
\begin{tabular}{lrrrp{0.1cm}lrrr}
\cmidrule(lr){1-4} \cmidrule(lr){6-9}
dataset & inst. & attr. & cls. & & dataset & inst. & attr. & cls. \\
\cmidrule(lr){1-4} \cmidrule(lr){6-9}
cardiotocography & 2126 & 21 & 10 & & letter-recognition & 20000 & 16 & 26 \\
diabetes & 768 & 8 & 2 & & optdigits & 5620 & 64 & 10\\
ecoli & 336 & 7 & 8 & & page-blocks & 5473 & 10 & 5 \\
glass & 214 & 9 & 7 & & parkinson & 195 & 22 & 2 \\
heart-statlog & 270 & 13 & 2 & & segment & 2310 & 18 & 7 \\
ionosphere & 351 & 34 & 2 & & spambase & 4601 & 57 & 2 \\
iris & 150 & 4 & 3 & & wine & 178 & 13 & 3 \\
landsat & 4435 & 36 & 6 & & & & & \\
\cmidrule(lr){1-4} \cmidrule(lr){6-9}
\end{tabular}
\end{center}
\caption{Datasets used in Experiment II, with number of instances, attributes and classes.}
\label{t:datasets}
\end{table}

The classifiers used are a KNN classifier with K$=3$ and an SVM classifier. We used the Weka implementation of SVM with the standard parameterisation:
{\it c} $= 1.0$, {\it tolerance parameter} $= 0.001$, {\it epsilon} $= 10^{-12}$, using a polynomial
kernel and a multinomial logistic regression model, with a ridge estimator as
calibrator. The only difference with the experiments of~\cite{carbonera2018efficient} is that we used
Weka 3.9.6 with the Python Weka wrapper, while they used Weka 3.8.

As discussed in Section~\ref{sec:soa}, the idea of the CNN~\cite{hart1968condensed} instance selection method is to find a subset of the original dataset that can correctly classify all the instances in the original dataset. Following this idea, Theorem~\ref{th:pjr-knn} suggests setting $t=2n$ for a KNN classifier with $\textrm{K}= 3$. Thus, for the UTCS rules, we computed reduced training sets for $\frac{t}{n}= 2, 1.5, 1, 0.75, 0.5,$ and $0.25$. For SeqP, we cannot set $t > n$, so we computed reduced training sets with SeqP and values for $\frac{t}{n}$ from $0.9$ to $0.1$ in steps of $0.1$.

Theorem~\ref{th:pjr-knn} also suggest that, for a KNN classifier with $\textrm{K}= 3$, it would be enough to employ a voting rule that satisfies $2$-PJR. In this experiment, we also used a variant of SEJR that we have called {\it Simple 2-EJR} (S$2$EJR). The only difference between SEJR and S$2$EJR is that, in S$2$EJR, we use a modified definition of EJR+ demand. In this modified definition, condition (iii) (see Definition~\ref{def:ejr+demand}) is (iii) $|A_i \cap W| < \min\{pl(c,W),2\}$ for each voter $v_i$ in $N^*$. In S$2$EJR, any voter with two of their preferred candidates in the winning set is considered {\it satisfied}. Therefore, such voters do not participate in the computation of a candidate's (modified) EJR+ demand. S$2$EJR fails EJR but satisfies $2$-EJR (and thus, $2$-PJR). In return, it achieves much higher reductions for a fixed $\frac{t}{n}$.

\begin{lemma}
  The rule S$2$EJR satisfies $2$-EJR.
\end{lemma}

The proof can be found in the appendixy.

We also compare the results obtained with our voting rules with those obtained with the original datasets (i.e. without any reduction) and with two baselines. The first baseline randomly selects some instances from the training set. We perform random reductions of the initial training set from $0.1$ to $0.9$ in steps of $0.1$ (we call this baseline `RANDOM'). The second baseline consists in removing those instances that do not belong to the local set of any other instance (we call this baseline `NOAPPROVED').

\begin{figure*}[thb]
    \centering
    \begin{subfigure}{0.49\textwidth}
      \includegraphics[width=\textwidth]{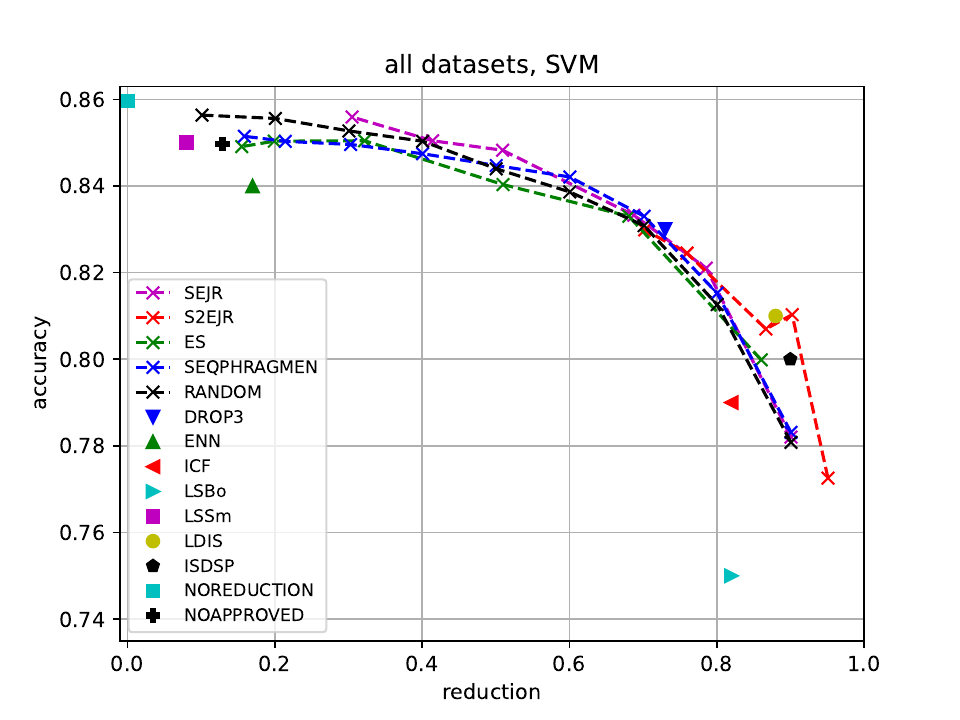}
    \end{subfigure}
    \hfill
    \begin{subfigure}{0.49\textwidth}
      \includegraphics[width=\textwidth]{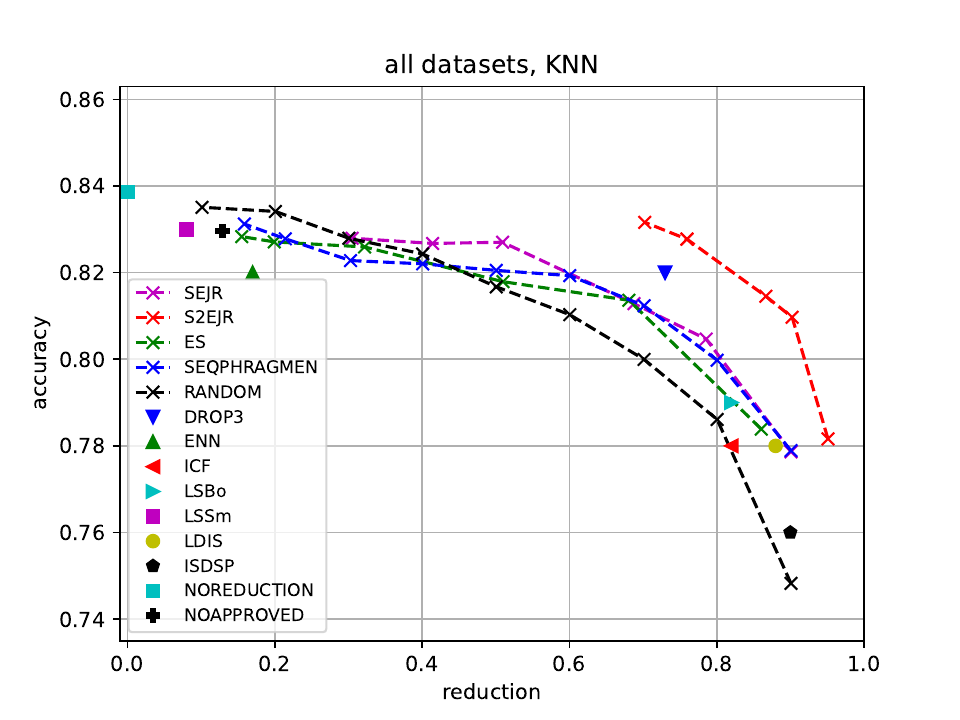}
    \end{subfigure}
    \caption{Average values of accuracy vs. reduction for SVM (left)
      and KNN (right).}
      \label{fig:ar-all-inst-excl}
\end{figure*}

Figure~\ref{fig:ar-all-inst-excl} shows average values of accuracy vs. reduction over the 15 datasets for SEJR, ES, SeqP and S$2$EJR (with the values of $\frac{t}{n}$ mentioned before), for the RANDOM baseline with reductions from $0.9$ to $0.1$, and for NOREDUCTION, NOAPPROVED and all instance selection algorithms considered by~\citet{carbonera2018efficient}.

Our first observation is that these figures are similar to Figure~2 by~\citet{cui2022dc}.\footnote{Figure~2 by~\citet{cui2022dc} considers in the horizontal coordinate the number of instances per class, while we use the value of reduction. Therefore, our figures are reversed compared to it.} Similarly to what they report, for state-of-the-art core set and data condensation algorithms, it can be observed that our rules perform better, but only by a small amount, than the RANDOM baseline (an exception to this is the rule S$2$EJR for KNN, which largely outperforms the RANDOM baseline).

We now focus on some particular $\frac{t}{n}$ values. Following the tradition of the instance selection literature, we focus on values of $\frac{t}{n}$ that achieve low reduction and values of $\frac{t}{n}$ that achieve high reduction (up to $90\%$, which is the maximum reduction reported by~\citet{carbonera2018efficient}). Therefore, we selected SEJR-$2$ and $0.25$, ES-$2$ and $0.25$, SeqP-$0.9$ and $0.1$, and S$2$EJR-$2$ and $0.5$.\footnote{We selected S$2$EJR-$0.5$ instead of S$2$EJR-$0.25$ because the average reduction of S$2$EJR-$0.25$ is above the limit of $90\%$.} We considered also RANDOM-$0.9$ and $0.1$, NOREDUCTION, NOAPPROVED, and all the algorithms considered by~\citet{carbonera2018efficient}.

Tables~\ref{t:inst-excl-rules} and~\ref{t:carbonera} show average values of accuracy for SVM and KNN and reduction obtained for all these algorithms (we consider a rule and a particular value of $\frac{t}{n}$ as an individual instance selection algorithm, and similarly for RANDOM and a specific fraction of the original dataset). We also include the values reported by~\citet{carbonera2018efficient} for comparison. In each row, we mark in boldface the algorithms that perform better. In the appendix, we provide tables with the values of accuracy and reduction for each separate dataset and algorithm.

A first criterion to compare algorithms would be Pareto efficiency with respect to the average values of accuracy and reduction. We say that algorithm $A$ is {\it better} than algorithm $B$ according to this criterion for a specific classifier if algorithm $A$ is not worse than algorithm $B$  with respect to accuracy and reduction, and $A$ is better than $B$ at least for one of these metrics. For instance, for SVM, our results show that SEJR-$2$ is better than ES-$2$, SeqP-$0.9$, ENN, LSSm, NOREDUCTION, NOAPPROVED and RANDOM-$0.9$ according to Pareto efficiency with respect to the average values of accuracy and reduction.

Considering this criterion, we can derive the following results from Tables~\ref{t:inst-excl-rules} and~\ref{t:carbonera}. For SVM, of all the baselines and state-of-the-art algorithms considered in Table~\ref{t:carbonera}, the only one that is not worse than any of the algorithms in Table~\ref{t:inst-excl-rules} is DROP3.
SEJR-$2$, ES-$2$, SeqP-$0.9$, and S$2$EJR-$0.5$ are not worse than any of the algorithms in Table~\ref{t:carbonera}.

\begin{table}[!thb]
\begin{center}
\setlength{\tabcolsep}{2pt}
\begin{tabular}{lrrrrrrrr}
\toprule
Average & SEJR-$2$ & S$2$EJR-$2$ & ES-$2$ & SeqP-$0.9$ & SEJR-$0.25$ & S$2$EJR-$0.5$ & ES-$0.25$ & SeqP-$0.1$\\
\midrule
SVM & ${\bf 0.86}$ & $0.83$ & $0.85$ & $0.85$ & $0.78$ & $0.81$ & $0.80$ & $0.78$\\
KNN & $0.83$ & $0.83$ & $0.83$ & $0.83$ & $0.78$ & $0.81$ & $0.78$ & $0.78$ \\
Red. & $0.31$ & $0.70$ & $0.16$ & $0.16$ & ${\bf 0.90}$ & ${\bf 0.90}$ & $0.86$ & ${\bf 0.90}$ \\
\bottomrule
\end{tabular}
\end{center}
\caption{Average accuracy for SVM and KNN and reduction results obtained with the voting rules considered and several values of $\frac{t}{n}$}
\label{t:inst-excl-rules}
\end{table}

\begin{table}[thb]
\begin{center}
\setlength{\tabcolsep}{2pt}
\begin{tabular}{lrrrrrrrrrrr}
\toprule
Average&DROP3&ENN&ICF&LSBo&LSSm&LDIS&ISDSP&NoR & NoA & R-$0.9$ & R-$0.1$ \\
\midrule
SVM&$0.83$&$0.84$&$0.79$&$0.75$&$0.85$&$0.81$&$0.80$& ${\bf 0.86}$&$0.85$&${\bf 0.86}$&$0.78$ \\
KNN&$0.82$&$0.82$&$0.78$&$0.79$&${ 0.83}$&$0.78$&$0.76$&${\bf 0.84}$&$0.83$&${\bf 0.84}$&$0.75$\\
Red.&$0.73$&$0.17$&$0.82$&$0.82$&$0.08$&$0.88$&${\bf 0.90}$&$0.00$&$0.13$&$0.10$&${\bf 0.90}$\\
\bottomrule
\end{tabular}
\end{center}
\caption{Average accuracy for SVM and KNN and reduction results reported by~\protect\citet{carbonera2018efficient} and obtained with several baselines (NoR means NOREDUCTION, NoA means NOAPPROVED and R-$0.9$ and $0.1$ means RANDOM-$0.9$ and $0.1$)}
\label{t:carbonera}
\end{table}

For KNN, of all the baselines and state-of-the-art algorithms considered in Table~\ref{t:carbonera}, again DROP3 plus NOREDUCTION and RANDOM-$0.9$ are not worse than any of the algorithms in Table~\ref{t:inst-excl-rules}. Of all the algorithms in Table~\ref{t:inst-excl-rules}, only ES-$0.25$ is worse than one of the algorithms in Table~\ref{t:carbonera}. It is worse than LDIS, which has the same accuracy but better reduction. All the other algorithms in Table~\ref{t:inst-excl-rules} are not worse than those in Table~\ref{t:carbonera}.

\begin{table}[!thb]
\begin{center}
\setlength{\tabcolsep}{2pt}
\begin{tabular}{|l|r|r|r|r|l|r|r|r|r|l|r|r|r|r|}
\hline
Algorithm & \multicolumn{2}{|c|}{SVM} & \multicolumn{2}{|c|}{KNN} &
Algorithm & \multicolumn{2}{|c|}{SVM} & \multicolumn{2}{|c|}{KNN} &
Algorithm & \multicolumn{2}{|c|}{SVM} & \multicolumn{2}{|c|}{KNN} \\ \cline{2-5} \cline{7-10} \cline{12-15}
 &$>$&$<$&$>$&$<$ &  &$>$&$<$&$>$&$<$ & &$>$&$<$&$>$&$<$ \\ \hline
SEJR-$2$ & ${\bf 5}$ & $0$& $3$& $0$ &
S$2$EJR-$2$ & $0$& $1$& $3$& $0$ &
ES-$2$ & $2$&$0$ & $2$& $0$ \\ \hline
SeqP-$0.9$ & $2$&$0$ & $2$& $0$ &
SEJR-$0.25$ & $1$& $1$& $4$& $0$ &
S$2$EJR-$0.5$ & ${\bf 5}$& $0$& ${\bf 5}$& $0$\\ \hline
ES-$0.25$ & $2$& $2$& $1$& $1$ &
SeqP-$0.1$ & $1$& $1$& $4$& $0$ &
DROP3 & $0$& $0$& $1$& $0$ \\ \hline
ENN & $0$& $0$&$0$ & $1$&
ICF & $1$& $2$& $0$&$2$ &
LSBo & $0$& $4$& $1$&$0$ \\ \hline
LSSm & $0$& $2$& $0$& $2$ &
LDIS & $2$& $0$& $1$& $0$ &
ISDSP & $3$& $0$& $1$& $0$\\ \hline
NoR & $0$& $1$& $0$& $1$ &
NoA & $1$& $0$& $1$& $0$ &
R-$0.9$ & $2$& $0$& $2$& $0$ \\ \hline
R-$0.1$ & $1$& $1$& $0$& $1$ & \multicolumn{10}{|c|}{ } \\ \hline
\end{tabular}
\end{center}
\caption{Comparison of each of the algorithms in Tables~\ref{t:inst-excl-rules} and~\ref{t:carbonera} with the baselines and state of the art algorithms in Table~\ref{t:carbonera} according to Pareto efficiency with respect to the average values of accuracy and reduction}
\label{t:comparison1}
\end{table}

We also count, for each of the algorithms considered in Tables~\ref{t:inst-excl-rules} and~\ref{t:carbonera}, the number of baselines and state-of-the-art algorithms considered in Table~\ref{t:carbonera} that are better and worse than the algorithm according to this criterion. Results are shown in Table~\ref{t:comparison1}. Columns labelled as $>$ indicate the number of algorithms in Table~\ref{t:carbonera} that are worse than the given algorithm, and columns labelled as $<$ indicate the number of algorithms in Table~\ref{t:carbonera} that are better than the given algorithm.

\subsection{Statistical analysis}

In this experiment, we cannot use confidence intervals to compare with state-of-the-art algorithms, because they have not been provided by~\citet{carbonera2018efficient}. Instead, we have carried out a statistical analysis of the results obtained in our experiments using the tests of~\citet{friedman1937use} and~\citet{bergmann1988improvements}, following the methodology recommended by~\citet{demvsar2006statistical} and~\citet{garcia2008extension}, and used to evaluate instance selection methods by~\citet{leyva2015three}.

The statistical analysis carried out in this experiment is based on building, for each dataset, a ranking of the algorithms according to the values for a given metric (accuracy for KNN or SVM, or reduction) obtained by each of them for that dataset.

Friedman tests address the following statistical problem. We want to compare the accuracy or the reduction of a set of algorithms and know whether we can reject the null hypothesis that all the algorithms considered are equivalent.

If the Friedman test allows for rejecting the null hypothesis, a post-hoc test can be conducted to compare each pair of algorithms. Following the methodology recommended by~\citet{garcia2008extension}, for comparing the performance of each pair of algorithms for a given metric, we use the~\citet{bergmann1988improvements} test. It allows us to accept or reject the null hypothesis that two algorithms are equivalent according to a given metric (accuracy for KNN or SVM, or reduction). If the null hypothesis is rejected, then the algorithm with a better average ranking (over all datasets) is considered {\it statistically better} than the other algorithm for the considered metric (accuracy for KNN or SVM, or reduction).

\begin{table}[!thb]
\begin{center}
\setlength{\tabcolsep}{2pt}
\begin{tabular}{|l|r|r|r|r|l|r|r|r|r|l|r|r|r|r|}
\hline
Algorithm & \multicolumn{2}{|c|}{SVM} & \multicolumn{2}{|c|}{KNN} &
Algorithm & \multicolumn{2}{|c|}{SVM} & \multicolumn{2}{|c|}{KNN} &
Algorithm & \multicolumn{2}{|c|}{SVM} & \multicolumn{2}{|c|}{KNN} \\ \cline{2-5} \cline{7-10} \cline{12-15}
 &$>$&$<$&$>$&$<$ &  &$>$&$<$&$>$&$<$ & &$>$&$<$&$>$&$<$ \\ \hline
SEJR-$2$ & ${\bf 4}$ & $0$& $2$& $1$ &
S$2$EJR-$2$ & $2$& $0$& $4$& $0$ &
ES-$2$ & $0$&$1$ & $0$& $2$ \\ \hline
SeqP-$0.9$ & $1$&$1$ & $0$& $1$ &
SEJR-$0.25$ & $0$& $0$& $1$& $0$ &
S$2$EJR-$0.5$ & $ 2$& $0$& ${\bf 6}$& $0$\\ \hline
ES-$0.25$ & $1$& $0$& $1$& $0$ &
SeqP-$0.1$ & $0$& $0$& $1$& $0$ &
DROP3 & $0$& $0/1$& $2/3$& $0$ \\ \hline
ENN & $0/1$& $0/1$&$0$ & $2/3/4$&
ICF & $1/2$& $0$& $0/1$&$0$ &
LSBo & $0$& $0/1$& $4/5$&$0$ \\ \hline
LSSm & $0/1$& $0$& $0$& $1/2$ &
LDIS & $0/1$& $0$& $1/2$& $0$ &
ISDSP & $0$& $0$& $1$& $0$\\ \hline
NoR & $0$& $0$& $0$& $2$ &
NoA & $0$& $1/2$& $0$& $1/2$ &
R-$0.9$ & $0$& $0$& $0$& $1$ \\ \hline
R-$0.1$ & $0$& $0$& $0$& $1$ & \multicolumn{10}{|c|}{ } \\ \hline
\end{tabular}
\end{center}
\caption{Comparison of each of the algorithms in Tables~\ref{t:inst-excl-rules} and~\ref{t:carbonera} with the baselines and state of the art algorithms in Table~\ref{t:carbonera} using the statistical analysis}
\label{t:comparison2}
\end{table}

We used this methodology to compare all the algorithms in Table~\ref{t:inst-excl-rules} with all the algorithms in Table~\ref{t:carbonera}, both in accuracy and reduction.
In the study, we used the significance level $\alpha= 0.05$.
Similarly to what we did with the average values of accuracy and reduction, we compare two algorithms using Pareto efficiency with respect to accuracy and reduction. We say that algorithm $A$ is {\it statistically better} than algorithm $B$ for a specific classifier if algorithm $A$ is not statistically worse than algorithm $B$ in accuracy and reduction, and $A$ is statistically better than $B$ at least for one of these metrics. For instance, in the appendix it is shown that, for SVMs, SEJR-$2$ is statistically better than DROP3 in accuracy (because the test of Bergmann and Hommel has rejected the hypotheses that SEJR-$2$ and DROP3 are statistically equivalent) and that SEJR-$2$ and DROP3 are statistically equivalent in reduction (because for reduction, the test of Bergmann and Hommel accepts the hypotheses that SEJR-$2$ and DROP3 are equivalent). Therefore, according to this criterion, we say that, for SVM, SEJR-$2$ is statistically better than DROP3.

For SVM, of all the baselines and state-of-the-art algorithms considered in Table~\ref{t:carbonera}, the algorithms that are not statistically worse than any of the algorithms in Table~\ref{t:inst-excl-rules} are RANDOM-$0.9$, LSSm, and LDIS. Of all the algorithms in Table~\ref{t:inst-excl-rules} only ES-$2$ and SeqP-$0.9$ are statistically worse than one of the algorithms in Table~\ref{t:carbonera}.

For KNN, of all the baselines and state-of-the-art algorithms considered in Table~\ref{t:carbonera}, the algorithms that are not statistically worse than any of the algorithms in Table~\ref{t:inst-excl-rules} are DROP3, LSBo, and LDIS. Of all the algorithms in Table~\ref{t:inst-excl-rules}, only SEJR-$2$, ES-$2$, and SeqP-$0.9$ are statistically worse than one or two of the algorithms in Table~\ref{t:carbonera}. All the other algorithms in Table~\ref{t:inst-excl-rules} are not statistically worse than any of the algorithms in Table~\ref{t:carbonera}.

Also, similarly to what we did with the average values, we provide in Table~\ref{t:comparison2}, for each of the algorithms considered in Tables~\ref{t:inst-excl-rules} and~\ref{t:carbonera}, the number of baselines and state of the art algorithms considered in Table~\ref{t:carbonera} that are statistically better and statistically worse than the algorithm. Columns labelled as $>$ (respectively, columns labelled as $<$) indicate the number of algorithms in Table~\ref{t:carbonera} that are statistically worse (respectively, that are statistically better) than the given algorithm.

The test of Bergmann and Hommel is computationally expensive, and we have only been able to compare up to $13$ algorithms with this test. To cope with this, we performed the statistical analysis 12 times. Each statistical analysis considers the $11$ algorithms in Table~\ref{t:carbonera}, plus the algorithms included in Table~\ref{t:inst-excl-rules} for one of the four rules considered: SEJR, ES, SeqP, and S$2$EJR. For each of these $4$ combinations of $13$ algorithms, we performed the statistical analysis for accuracy with SVM, accuracy with KNN, and reduction.
A drawback of this approach is that each pair of algorithms in Table~\ref{t:carbonera} is compared 4 times for each type of analysis. In some rare cases (around $3\%$ of the comparisons), the test of Bergmann and Hommel gives different results in some of the repeated comparisons, affecting the values in Table~\ref{t:comparison2}. In such cases, we present in Table~\ref{t:comparison2} the possible values depending on which of the repeated comparisons is chosen.
Detailed descriptions of the methodology and the results of the statistical analysis are provided in the appendix.

\section{Conclusions}

In this paper, we establish a connection between the problem of finding a representative subset of a dataset (CS/I selection) in machine learning and notions of representation from the computational social choice domain. Among the strengths of our approach, we can highlight the following. First of all, our approach is not a heuristic, but grounded on well established principles of proportional representation, recently developed by the computational social choice community. Secondly, a significant strength of this work is the rigorous evaluation conducted. We have evaluated our approach using deep learning classifiers and classic machine learning classifiers (KNN and SVM), and with several standard datasets, and we have compared our results with several prominent core set selection and instance selection algorithms available in the state of the art. Finally, we have assessed the accuracy obtained in our experiments using appropriate statistical methods (confidence intervals in the neural networks experiment and Friedman and Bergmann-Hommel tests in the experiment with classic machine learning classifiers).

In the first experiment, we compared the accuracy obtained with our rules (ES, SEJR, and SeqP) with several state-of-the-art core set algorithms and the random baseline. Remarkably, the accuracy obtained with our rules was better than that of the state-of-the-art core set algorithms in all cases, with statistically significant differences in all cases but one, in which herding achieved a similar accuracy to ES.

\citet{rebuffi2017icarl} discuss that one advantage of using herding for class incremental learning systems like iCaRL is the fact that instances are ordered. Then, when a new class enters the system, if we need to reduce the number of instances stored per class, we do not need to run the core set selection algorithm again. It is sufficient to remove the last instances (in the order given by herding) stored for each of the old classes. In computational social choice, the equivalent property is committee monotonicity~\citep{elkind:scw17}. Informally speaking, if a multi-winner voting rule satisfies committee monotonicity, then for each election, the committee obtained when we subtract one from the target committee size is a subset of the original committee. ES and SEJR do not satisfy committee monotonicity, but SeqP does. Similarly to herding, SeqP outputs an ordered sequence of instances for each class, so removing an instance from each class consists of removing the last instance of each sequence. We believe that class incremental learning systems like iCaRL could benefit from replacing herding with SeqP.

In the second experiment, we compared our rules with several state-of-the-art instance selection algorithms and baselines, using two classic machine learning classifiers: KNN and SVM. We observe that instance selection methods (except ISDSP) cannot be targeted to a particular number of instances per class. Therefore, they could not have been employed in the first experiment. Since each algorithm obtains different values for accuracy and reduction, in this case we adopted an approach based on Pareto efficiency to compare the algorithms.

The statistical analysis of the results of this second experiment is based on the rankings of the algorithms obtained for each dataset. Compared to the statistical analysis based on the computation of confidence intervals, this approach has advantages and disadvantages. The main advantage is that the analysis based on rankings does not make any assumption about the normality of the distributions and, therefore, the results obtained are more robust. In contrast, the test of Bergmann and Hommel is not very powerful and, thus, it accepts many hypotheses of equivalence of pairs of classifiers. Another problem of the tests based on rankings is that they do not consider the differences in the values obtained with the algorithms, only the rankings. For instance, in all the datasets, the reduction obtained with S$2$EJR-$2$ was larger than that obtained with ENN, with differences between $0.20$ and $0.86$. However, when building rankings with the state-of-the-art algorithms, the baselines, and S$2$EJR-$2$ and $0.5$, it can be observed that ENN and S$2$EJR-$2$ typically appear consecutive in the reduction rankings. It is difficult to believe that S$2$EJR-$2$ is not better than ENN in reduction, but the test of Bergmann and Hommel says that such hypotheses cannot be discarded.

The statistical analysis performed in the second experiment is, therefore, less conclusive.
However, it is remarkable that both in the analysis based on average values (Table~\ref{t:comparison1}) and in the statistical analysis (Table~\ref{t:comparison2}), the best algorithm for SVM was SEJR-$2$ and the best one for KNN was S$2$EJR-$0.5$.
We believe that all the analyses performed, and in particular Figure~\ref{fig:ar-all-inst-excl}, show that for KNN with K$= 3$, when a high reduction is desired, S$2$EJR outperforms all the state-of-the-art algorithms and baselines.

Although all the voting rules considered in this paper can be executed in polynomial time, their theoretical time complexity is $n$ times slower than that of herding and $k$-Center. In contrast, the time complexity of state-of-the-art instance selection methods that use local sets would be comparable to ours, since computing local sets requires computing distances between all instances in the dataset. Therefore, the time needed to compute the local sets grows quadratically with the dataset size. We also note that the computation of all our voting rules is easily parallelizable, allowing us to significantly reduce the computation time when sufficient computational resources are available.

The theoretical result we have obtained for KNN classifiers, along with the evaluation performed in the experiments, demonstrate the robustness and performance of our approach. However, we believe it would be convenient to investigate further the possible limitations of this approach. For instance, imagine a dataset in which the instances that belong to a class are divided into two regions, with most of the instances placed in one area and only a few in the other. If the target committee is small enough, it may be possible that all instances from the less densely populated area are removed in the reduced dataset. We can fix this by increasing the target committee size, but that would increase the size of the reduced dataset. A possible solution for this problem could be to run separate elections for the two areas, with different target committee sizes.

Several interesting lines of continuation of this work can be outlined. We have established a theoretical result that justifies the use of voting rules that satisfy the $\frac{K+1}{2}$-PJR to reduce a dataset to be used with a KNN classifier. It would be interesting to obtain similar results for other types of classifiers, like SVMs using a certain Kernel or different neural network architectures. We note that the computational social choice community has developed a rich set of proportionality axioms (we refer to the survey by~\citet{multiwinnerApprovalSurvey} for a discussion on many of these axioms, although the field is quite active and new axioms have been proposed since the publication of this survey, like EJR+~\citep{brill2023robust} and FPJR~\citep{kalayci2025full}). Therefore, the notion of proportionality that best fits each classifier could vary from case to case.

The experiments that we have presented in this paper use standard publicly available datasets. For such standard datasets it is natural to expect that the presence of noisy instances would be quite low. It would be interesting to evaluate our approach with datasets that contain a significant number of noisy instances.

Finally, we believe that the connection that we have established between notions of representation in the (computational) social choice domain and the concept of representation in machine learning could be applied to other applications. For instance, we believe that the social choice model that we have created for the instance selection problem could be adapted to evaluate whether a synthetic dataset is representative of a real dataset from which the synthetic dataset was generated.

\section*{Acknowledgments}

This work has been partially funded by the ``Generation of Reliable
Synthetic Health Data for Federated Learning in Secure Data Spaces''
Research Project (PID2022-141045OB-C43 (AEI/ERDF, EU)) funded by
MICIU/AEI/ 10.13039/501100011033 and by ``ERDF A way of making Europe''
by the ``European Union''.

\bibliographystyle{named}
\bibliography{IS,dhondt}

\appendix

\section{Summary}
\label{app:summary}

This appendix contains the following material.

First of all, in Section~\ref{app:es_seqp}, we detail the operation of the rules ES and SeqP, including a pseudocode to compute the outcome of both rules, and analyse the time and memory complexity of those algorithms. We also include an analysis of the memory complexity of the rule SEJR, based on a pseudocode of SEJR given in~\citep{SEL17a}. In Section~\ref{app:s2ejr} we prove some results for S$2$EJR that where omitted in the main paper.

Tables~\ref{t:acc-bigk-excl-svm} to~\ref{t:baselines-red} contain the values of accuracy (for KNN and SVM) and reduction obtained for each dataset for the algorithms included in the main paper in Table~3 and the baselines included in the main paper in Table~4. For comparison, we also include the results reported by~\citet{carbonera2018efficient} for the state-of-the-art instance selection algorithms they considered in Tables~\ref{t:carbonera-SVM} to~\ref{t:carbonera-red}.

In the experiment with neural networks, the reduced datasets must contain a fixed number (10 or 50) of instances per class. However, as discussed in the main paper, the rules SEJR and ES, given a specific target committee size $t$, usually return a committee composed of fewer than $t$ winners (the instances that will be included in the reduced dataset). In the main paper, we addressed this problem following the approach suggested in~\citep{peters2020proportionality}. There, the authors first suggest obtaining the committee output by ES and then applying the rule SeqP to add additional winners until the committee size is $t$. An alternative approach could be to increase the target committee size $t$ as much as possible, so that the size of the committee output by SEJR or ES is as close as possible (but not greater than) the target number of instances per class, and then run again SeqP to complete the committee if necessary. Of course, this search for the optimal target committee size would require multiple runs of SEJR or ES, which would imply a significant increase in the time needed to run the algorithms. The main advantage of this approach is to differentiate SEJR and ES from SeqP as much as possible. In Table~\ref{tab:exp-1-accuracyV2} we report the results obtained in the experiment with neural networks with this alternative for SEJR and ES. In~Table~\ref{tab:exp-1-accuracyV2} we report the accuracy and confidence intervals obtained in the experiment with neural networks. The values for SEJR and ES have been obtained by optimizing the target committee size, while the values of SeqP, the state-of-the-art baselines and the Random baseline are identical to those reported in the main paper. 

Comparing the results obtained for SEJR and ES with or without target committee size optimization, we can see that, as expected, the values obtained with target committee size optimization are more different from those of SeqP than the values obtained without target committee size optimization. However, they are not necessarily better, with the exception of CIFAR-10, with 10 instances per class, in which the accuracy values obtained with target committee size optimization are better in all cases except one.

Finally, in Section~\ref{sec:suppl_statistic}, we detail the process that we have followed in the statistical analysis of Experiment II and show its results with more detail.

\begin{table}[!thb]
\begin{center}
\setlength{\tabcolsep}{2pt}
\begin{tabular}{lrrrr}
\toprule
dataset & SEJR-2 & S2EJR-2 & ES-2 & SeqP-0.9 \\
\midrule
cardiotocography & 0.77 & 0.76 & 0.76 & 0.77 \\
diabetes & {\bf 0.77} & 0.76 & 0.76 & 0.76 \\
ecoli & 0.82 & 0.83 & 0.82 & 0.80 \\
glass & 0.54 & 0.49 & 0.47 & 0.54 \\
heart-statlog & 0.83 & 0.81 & 0.82 & 0.83 \\
ionosphere & 0.85 & 0.78 & 0.86 & 0.85 \\
iris & {\bf 0.97} & 0.89 & 0.96 & 0.95 \\
landsat & {\bf 0.87} & 0.86 & {\bf 0.87} & {\bf 0.87} \\
letter-recognition & 0.82 & 0.75 & 0.82 & 0.82 \\
optdigits & 0.98 & 0.97 & 0.98 & 0.98 \\
page-blocks & {\bf 0.94} & 0.93 & {\bf 0.94} & {\bf 0.94} \\
parkinson & 0.87 & 0.83 & 0.87 & 0.86 \\
segment & {\bf 0.93} & 0.92 & {\bf 0.93} & 0.92 \\
spambase & {\bf 0.92} & 0.90 & 0.91 & 0.91 \\
wine & 0.97 & 0.97 & 0.96 & 0.96 \\
\midrule
average & {\bf 0.86} & 0.83 & 0.85 & 0.85 \\
\bottomrule
\end{tabular}
\end{center}
\caption{Accuracy results obtained in Experiment II with the voting rules considered for big values of $\frac{t}{n}$ with an SVM classifier}
\label{t:acc-bigk-excl-svm}
\end{table}

\begin{table}[!thb]
\begin{center}
\setlength{\tabcolsep}{2pt}
\begin{tabular}{lrrrr}
\toprule
dataset & SEJR-2 & S2EJR-2 & ES-2 & SeqP-0.9 \\
\midrule
cardiotocography & 0.69 & 0.69 & 0.70 & 0.70 \\
diabetes & 0.69 & 0.71 & 0.68 & 0.68 \\
ecoli & 0.86 & 0.86 & {\bf 0.87} & {\bf 0.87} \\
glass & 0.64 & 0.67 & 0.64 & 0.67 \\
heart-statlog & 0.66 & 0.66 & 0.66 & 0.66 \\
ionosphere & 0.85 & 0.87 & 0.84 & 0.83 \\
iris & {\bf 0.97} & {\bf 0.97} & {\bf 0.97} & {\bf 0.97} \\
landsat & 0.89 & 0.89 & 0.90 & 0.90 \\
letter-recognition & 0.95 & 0.94 & 0.95 & 0.95 \\
optdigits & 0.98 & 0.98 & {\bf 0.99} & 0.98 \\
page-blocks & {\bf 0.96} & 0.95 & {\bf 0.96} & {\bf 0.96} \\
parkinson & 0.83 & 0.83 & 0.83 & 0.84 \\
segment & 0.94 & 0.94 & 0.95 & 0.94 \\
spambase & 0.79 & 0.79 & 0.80 & 0.81 \\
wine & 0.72 & 0.71 & 0.68 & 0.71 \\
\midrule
average & 0.83 & 0.83 & 0.83 & 0.83 \\
\bottomrule
\end{tabular}
\end{center}
\caption{Accuracy results obtained in Experiment II with the voting rules considered for big values of $\frac{t}{n}$ with a KNN classifier}
\label{t:acc-bigk-excl-knn}
\end{table}

\begin{table}[!thb]
\begin{center}
\setlength{\tabcolsep}{2pt}
\begin{tabular}{lrrrr}
\toprule
dataset & SEJR-2 & S2EJR-2 & ES-2 & SeqP-0.9 \\
\midrule
cardiotocography & 0.32 & 0.52 & 0.23 & 0.19 \\
diabetes & 0.44 & 0.58 & 0.31 & 0.24 \\
ecoli & 0.39 & 0.67 & 0.17 & 0.14 \\
glass & 0.32 & 0.56 & 0.24 & 0.22 \\
heart-statlog & 0.51 & 0.59 & 0.42 & 0.33 \\
ionosphere & 0.36 & 0.81 & 0.21 & 0.20 \\
iris & 0.15 & 0.82 & 0.02 & 0.10 \\
landsat & 0.31 & 0.78 & 0.09 & 0.10 \\
letter-recognition & 0.24 & 0.75 & 0.04 & 0.10 \\
optdigits & 0.34 & 0.88 & 0.01 & 0.10 \\
page-blocks & 0.17 & 0.90 & 0.04 & 0.10 \\
parkinson & 0.23 & 0.63 & 0.13 & 0.11 \\
segment & 0.17 & 0.79 & 0.03 & 0.10 \\
spambase & 0.34 & 0.62 & 0.19 & 0.14 \\
wine & 0.29 & 0.63 & 0.23 & 0.21 \\
\midrule
average & 0.31 & 0.70 & 0.16 & 0.16 \\
\bottomrule
\end{tabular}
\end{center}
\caption{Reduction results obtained in Experiment II with the voting rules considered for big values of $\frac{t}{n}$}
\label{t:red-bigk-excl}
\end{table}

\begin{table}[!thb]
\begin{center}
\setlength{\tabcolsep}{2pt}
\begin{tabular}{lrrrr}
\toprule
dataset & SEJR-0.25 & S2EJR-0.5 & ES-0.25 & SeqP-0.1 \\
\midrule
cardiotocography & 0.62 & 0.70 & 0.64 & 0.66 \\
diabetes & 0.76 & 0.75 & {\bf 0.77} & 0.76 \\
ecoli & 0.77 & 0.79 & 0.77 & 0.76 \\
glass & 0.46 & 0.49 & 0.46 & 0.50 \\
heart-statlog & 0.69 & 0.75 & 0.74 & 0.75 \\
ionosphere & 0.83 & 0.79 & 0.84 & 0.83 \\
iris & 0.86 & 0.89 & 0.91 & 0.80 \\
landsat & 0.85 & 0.85 & 0.85 & 0.85 \\
letter-recognition & 0.75 & 0.74 & 0.77 & 0.73 \\
optdigits & 0.95 & 0.96 & 0.97 & 0.96 \\
page-blocks & 0.93 & 0.93 & 0.94 & 0.93 \\
parkinson & 0.82 & 0.86 & 0.85 & 0.78 \\
segment & 0.87 & 0.90 & 0.88 & 0.85 \\
spambase & 0.90 & 0.91 & 0.89 & 0.89 \\
wine & 0.65 & 0.85 & 0.72 & 0.70 \\
\midrule
average & 0.78 & 0.81 & 0.80 & 0.78 \\
\bottomrule
\end{tabular}
\end{center}
\caption{Accuracy results obtained in Experiment II with the voting rules considered for small values of $\frac{t}{n}$ with an SVM classifier}
\label{t:acc-smallk-excl-svm}
\end{table}

\begin{table}[!thb]
\begin{center}
\setlength{\tabcolsep}{2pt}
\begin{tabular}{lrrrr}
\toprule
dataset & SEJR-0.25 & S2EJR-0.5 & ES-0.25 & SeqP-0.1 \\
\midrule
cardiotocography & 0.52 & 0.61 & 0.53 & 0.56 \\
diabetes & 0.71 & 0.74 & 0.72 & 0.71 \\
ecoli & 0.78 & 0.84 & 0.78 & 0.77 \\
glass & 0.55 & 0.61 & 0.56 & 0.59 \\
heart-statlog & 0.63 & 0.64 & 0.62 & 0.64 \\
ionosphere & 0.84 & 0.86 & 0.84 & 0.84 \\
iris & 0.93 & {\bf 0.97} & 0.91 & 0.93 \\
landsat & 0.87 & 0.88 & 0.87 & 0.86 \\
letter-recognition & 0.88 & 0.92 & 0.89 & 0.84 \\
optdigits & 0.96 & 0.98 & 0.98 & 0.96 \\
page-blocks & 0.94 & 0.95 & 0.94 & 0.93 \\
parkinson & 0.80 & 0.80 & 0.80 & 0.78 \\
segment & 0.86 & 0.90 & 0.88 & 0.83 \\
spambase & 0.75 & 0.76 & 0.74 & 0.75 \\
wine & 0.68 & 0.70 & 0.69 & 0.69 \\
\midrule
average & 0.78 & 0.81 & 0.78 & 0.78 \\
\bottomrule
\end{tabular}
\end{center}
\caption{Accuracy results obtained in Experiment II with the voting rules considered for small values of $\frac{t}{n}$ with a KNN classifier}
\label{t:acc-smallk-excl-knn}
\end{table}

\begin{table}[!thb]
\begin{center}
\setlength{\tabcolsep}{2pt}
\begin{tabular}{lrrrr}
\toprule
dataset & SEJR-0.25 & S2EJR-0.5 & ES-0.25 & SeqP-0.1 \\
\midrule
cardiotocography & {\bf 0.95} & 0.87 & 0.93 & 0.90 \\
diabetes & {\bf 0.95} & 0.90 & 0.93 & 0.90 \\
ecoli & {\bf 0.93} & 0.88 & 0.88 & 0.90 \\
glass & {\bf 0.93} & 0.88 & 0.92 & 0.90 \\
heart-statlog & {\bf 0.97} & 0.92 & 0.96 & 0.90 \\
ionosphere & 0.87 & 0.91 & 0.83 & 0.90 \\
iris & 0.83 & {\bf 0.92} & 0.80 & 0.90 \\
landsat & 0.88 & {\bf 0.92} & 0.82 & 0.90 \\
letter-recognition & 0.87 & 0.88 & 0.81 & {\bf 0.90} \\
optdigits & 0.88 & {\bf 0.94} & 0.78 & 0.90 \\
page-blocks & 0.86 & 0.95 & 0.78 & 0.90 \\
parkinson & {\bf 0.90} & 0.86 & 0.87 & {\bf 0.90} \\
segment & 0.87 & {\bf 0.92} & 0.81 & 0.90 \\
spambase & {\bf 0.92} & 0.89 & 0.89 & 0.90 \\
wine & {\bf 0.91} & 0.90 & 0.90 & 0.90 \\
\midrule
average & {\bf 0.90} & {\bf 0.90} & 0.86 & {\bf 0.90} \\
\bottomrule
\end{tabular}
\end{center}
\caption{Reduction results obtained in Experiment II with the voting rules considered for small values of $\frac{t}{n}$}
\label{t:red-smallk-excl}
\end{table}

\begin{table}[!thb]
\begin{center}
\setlength{\tabcolsep}{2pt}
\begin{tabular}{lrrrr}
\toprule
dataset & NoR & NoA & R-0.9 &  R-0.1 \\
\midrule
cardiotocography & 0.78 & 0.77 & 0.78 & 0.65 \\
diabetes & {\bf 0.77} & 0.75 & 0.76  & 0.72 \\
ecoli & 0.83 & 0.82 & 0.82 & 0.71 \\
glass & 0.54 & 0.51 & 0.52 & 0.42 \\
heart-statlog & {\bf 0.85} & 0.82 & 0.83 & 0.77 \\
ionosphere & {\bf 0.88} & 0.84 & {\bf 0.88}  & 0.81 \\
iris & {\bf 0.97} & 0.96 & 0.96 & 0.72 \\
landsat & {\bf 0.87} & {\bf 0.87} & {\bf 0.87} & 0.84 \\
letter-recognition & 0.82 & 0.82 & 0.82 & 0.72 \\
optdigits & 0.98 & 0.98 & 0.98 & 0.96 \\
page-blocks & 0.93 & 0.93 & 0.93 & 0.92 \\
parkinson & 0.87 & 0.86 & 0.87 & 0.80 \\
segment & {\bf 0.93} & {\bf 0.93} & {\bf 0.93} & 0.87 \\
spambase & 0.90 & 0.91 & 0.90 & 0.88 \\
wine & 0.98 & 0.97 & {\bf 0.99} & 0.92 \\
\midrule
average & {\bf 0.86} & 0.85 & {\bf 0.86} & 0.78 \\
\bottomrule
\end{tabular}
\end{center}
\caption{Accuracy results obtained in Experiment II with the original datasets and the baselines and an SVM classifier}
\label{t:baselines-svm}
\end{table}

\begin{table}[!thb]
\begin{center}
\setlength{\tabcolsep}{2pt}
\begin{tabular}{lrrrr}
\toprule
dataset & NoR & NoA & R-0.9 & R-0.1 \\
\midrule
cardiotocography & {\bf 0.74} & 0.70 & 0.72 & 0.53 \\
diabetes & 0.69 & 0.69 & 0.69 & 0.68 \\
ecoli & 0.85 & {\bf 0.87} & 0.85 & 0.76 \\
glass & 0.69 & 0.65 & 0.68 & 0.50 \\
heart-statlog & 0.65 & 0.66 & 0.64 & 0.58 \\
ionosphere & 0.84 & 0.84 & 0.84 & 0.74 \\
iris & 0.96 & {\bf 0.97} & 0.96 & 0.86 \\
landsat & {\bf 0.91} & {\bf 0.91} & 0.90 & 0.86 \\
letter-recognition & {\bf 0.96} & 0.95 & {\bf 0.96} & 0.83 \\
optdigits & {\bf 0.99} & {\bf 0.99} & {\bf 0.99} & 0.96 \\
page-blocks & {\bf 0.96} & {\bf 0.96} & {\bf 0.96} & 0.94 \\
parkinson & 0.85 & 0.83 & 0.85 & 0.75 \\
segment & {\bf 0.96} & 0.95 & 0.95 & 0.83 \\
spambase & 0.81 & 0.80 & 0.81 & 0.71 \\
wine & 0.73 & 0.67 & 0.72 & 0.68 \\
\midrule
average & {\bf 0.84} & 0.83 & {\bf 0.84} & 0.75 \\
\bottomrule
\end{tabular}
\end{center}
\caption{Accuracy results obtained in Experiment II with the original datasets and the baselines and a KNN classifier}
\label{t:baselines-knn}
\end{table}

\begin{table}[!thb]
\begin{center}
\setlength{\tabcolsep}{2pt}
\begin{tabular}{lrrr}
\toprule
dataset & NoA & R-0.9 & R-0.1 \\
\midrule
cardiotocography & 0.19 & 0.10 & 0.90 \\
diabetes & 0.24 & 0.10 & 0.90 \\
ecoli & 0.13 & 0.10 & 0.90 \\
glass & 0.22 & 0.10 & 0.90 \\
heart-statlog & 0.34 & 0.10 & 0.90 \\
ionosphere & 0.20 & 0.10 & 0.90 \\
iris & 0.02 & 0.10 & 0.90 \\
landsat & 0.06 & 0.10 & 0.90 \\
letter-recognition & 0.02 & 0.10 & {\bf 0.90} \\
optdigits & 0.00 & 0.10 & 0.90 \\
page-blocks & 0.03 & 0.10 & 0.90 \\
parkinson & 0.11 & 0.10 & {\bf 0.90} \\
segment & 0.02 & 0.10 & 0.90 \\
spambase & 0.14 & 0.10 & 0.90 \\
wine & 0.21 & 0.10 & 0.90 \\
\midrule
average & 0.13 & 0.10 & {\bf 0.90} \\
\bottomrule
\end{tabular}
\end{center}
\caption{Reduction results obtained in Experiment II with the baselines}
\label{t:baselines-red}
\end{table}

\begin{table}[thb]
\begin{center}
\setlength{\tabcolsep}{2pt}
\begin{tabular}{lrrrrrrr}
\toprule
dataset&DROP3&ENN&ICF&LSBo&LSSm&LDIS&ISDSP \\
\midrule
ctg. &0.64&0.67&0.64&0.62&0.67&0.62&0.59 \\
diabetes&0.75&{\bf 0.77}&0.76&0.75&{\bf 0.77}&0.75&0.73 \\
ecoli&0.81&0.82&0.78&0.74&0.83&0.77&0.78 \\
glass&0.47&0.49&0.49&0.42&0.55&0.50&0.51 \\
heart-statlog&0.81&0.83&0.79&0.81&0.84&0.81&0.78 \\
ionosphere&0.81&0.87&0.58&0.45&{\bf 0.88}&0.84&0.86 \\
iris&0.94&0.96&0.73&0.47&0.96&0.81&0.80 \\
landsat&0.86&{\bf 0.87}&0.85&0.85&{\bf 0.87}&0.84&0.84 \\
letter &0.80&{\bf 0.84}&0.75&0.73&{\bf 0.84}&0.75&0.74 \\
optdigits&0.98&0.98&0.97&0.98&{\bf 0.99}&0.96&0.97 \\
page-blocks&0.93&{\bf 0.94}&0.93&0.92&{\bf 0.94}&{\bf 0.94}&0.91 \\
parkinson&0.85&0.87&0.85&0.82&0.87&0.82&0.85 \\
segment&0.91&0.92&0.91&0.80&0.91&0.89&0.88 \\
spambase&0.90&0.90&0.90&0.90&0.90&0.89&0.87 \\
wine&0.93&0.95&0.94&0.96&0.97&0.94&0.93 \\
\midrule
average&0.83&0.84&0.79&0.75&0.85&0.81&0.80 \\
\bottomrule
\end{tabular}
\end{center}
\caption{Accuracy results reported by~\protect\cite{carbonera2018efficient} with an SVM classifier}
\label{t:carbonera-SVM}
\end{table}

\begin{table}[!bth]
\begin{center}
\setlength{\tabcolsep}{2pt}
\begin{tabular}{lrrrrrrr}
\toprule
dataset&DROP3&ENN&ICF&LSBo&LSSm&LDIS&ISDSP \\
\midrule
ctg. &0.63&0.64&0.57&0.55&0.67&0.54&0.50 \\
diabetes&0.72&0.72&0.72&0.73&0.72&0.68&0.65 \\
ecoli&0.84&0.84&0.79&0.79&0.86&0.82&0.82 \\
glass&0.63&0.63&0.64&0.54&{\bf 0.71}&0.62&0.55 \\
heart-statlog&{\bf 0.67}&0.64&0.63&0.66&0.66&{\bf 0.67}&0.63 \\
ionosphere&0.82&0.83&0.82&{ 0.88}&0.86&0.85&0.85 \\
iris&{\bf 0.97}&{\bf 0.97}&0.95&0.95&0.96&0.95&0.95 \\
landsat&0.88&{ 0.90}&0.83&0.86&{ 0.90}&0.87&0.86 \\
letter &0.88&0.92&0.80&0.73&0.93&0.79&0.71 \\
optdigits&0.97&0.98&0.91&0.91&0.98&0.95&0.94 \\
page-blocks&0.95&{\bf 0.96}&0.93&0.94&{\bf 0.96}&0.94&0.77 \\
parkinson&0.86&{\bf 0.88}&0.83&0.85&0.85&0.74&0.79 \\
segment&0.92&0.94&0.87&0.83&0.94&0.88&0.89 \\
spambase&0.79&0.81&0.79&0.81&{\bf 0.82}&0.75&0.77 \\
wine&0.69&0.66&0.66&0.74&0.71&0.69&{ 0.75} \\
\midrule
average&0.82&0.82&0.78&0.79&{ 0.83}&0.78&0.76 \\
\bottomrule
\end{tabular}
\end{center}
\caption{Accuracy results reported by~\protect\cite{carbonera2018efficient} with a KNN classifier}
\label{t:carbonera-KNN}
\end{table}

\begin{table}[!thb]
\begin{center}
\setlength{\tabcolsep}{2pt}
\begin{tabular}{lrrrrrrr}
\toprule
dataset&DROP3&ENN&ICF&LSBo&LSSm&LDIS&ISDSP \\
\midrule
ctg.&0.70&0.32&0.71&0.69&0.14&0.86&0.90 \\
diabetes&0.77&0.31&0.85&0.76&0.13&0.90&0.90 \\
ecoli&0.72&0.17&0.87&0.83&0.09&0.92&0.90 \\
glass&0.75&0.35&0.69&0.70&0.13&0.90&0.90 \\
heart&0.74&0.35&0.78&0.67&0.15&0.93&0.90 \\
ionosphere&0.86&0.15&{\bf 0.96}&0.81&0.04&0.91&0.90 \\
iris&0.70&0.04&0.61&{\bf 0.92}&0.05&0.87&0.90 \\
landsat&0.72&0.10&0.91&0.88&0.05&{\bf 0.92}&{\bf 0.90} \\
letter&0.68&0.05&0.80&0.84&0.04&0.82&0.90 \\
optdigits&0.72&0.01&0.93&0.92&0.02&0.92&0.90 \\
page-blocks&0.71&0.04&0.95&{\bf 0.96}&0.03&0.87&0.90 \\
parkinson&0.72&0.15&0.80&0.87&0.11&0.83&{\bf 0.90} \\
segment&0.68&0.05&0.79&0.90&0.05&0.83&0.90 \\
spambase&0.74&0.19&0.79&0.82&0.10&0.82&0.90 \\
wine&0.80&0.30&0.82&0.75&0.11&0.88&0.90 \\
\midrule
average&0.73&0.17&0.82&0.82&0.08&0.88&{\bf 0.90} \\
\bottomrule
\end{tabular}
\end{center}
\caption{Reduction results reported by~\protect\cite{carbonera2018efficient}}
\label{t:carbonera-red}
\end{table}

\begin{algorithm}[!htb]
\caption{Pseudocode for computing the outcome of ES}\label{alg:es}
\noindent\makebox[\linewidth]{\rule{\textwidth}{0.4pt}}
{\bf Input}: an approval-based multi-winner election $(N, C, \mathcal{A}, t)$\\
{\bf Output}: the set of winners $W$\\
\noindent\makebox[\linewidth]{\rule{\textwidth}{0.4pt}}

\algblockdefx[ForEach]{ForEach}{EndForEach}[1]
{\textbf{foreach} #1 \textbf{do}}{\textbf{end foreach}} 
{\small
\begin{algorithmic}[1]
\State Compute the number of voters $n_c$ that approve of each candidate $c$
\State $W\gets \emptyset$
\State  $\ell \gets []$ \Comment{$\ell$ is a list that contains for each voter $v_i$ the pair $(\rho(v_i),i)$}
\ForEach{$v_i \in N$}
  \State $\ell.\textrm{append}(1,i)$
\EndForEach

\Repeat
  \State $\gamma \gets -1$
  \State Sort $\ell$ in increasing order of $\rho(v_i)$
  \ForEach{$c \in C \setminus W$}
    \State $x \gets 0$ \Comment{Number of voters that approve of candidate $c$ traversed}  
    \State $p \gets 0$ \Comment{Sum of the values of $\rho(v_i)$ of those $x$ voters}
    \For{$(\rho(v_i),i)$ \textbf{in} $\ell$}
       \If{$c \in A_i$}
        \If{$\frac{n}{t} - p \leq \rho(v_i) (n_c - x)$}
          \State $\gamma' \gets \frac{\frac{n}{t} - p}{n_c - x}$
          \If{$\gamma= -1$ \textbf{or} $\gamma > \gamma'$}
            \State $\gamma \gets \gamma'$
          \EndIf
          \BREAK
        \Else  
          \State $p \gets p + \rho(v_i)$; $x \gets x + 1$
        \EndIf
      \EndIf
    \EndFor
  \EndForEach
  \If{$\gamma > 0$}
    \State $W \gets W \cup \{w\}$
    \For{$j=1$ \textbf{to} $n$}
      \State $(\rho(v_i),i) \gets \ell[j]$ 
      \If{$w \in A_i$}
        \State $\ell[j]= (\max \{0, \rho(v_i) - \gamma\}, i)$
      \EndIf
    \EndFor 
  \EndIf
\Until{$\gamma = -1$}
\State \textbf{return} $W$
\end{algorithmic}
}
\end{algorithm}

\begin{algorithm}[!htb]
\caption{Pseudocode for computing the outcome of SeqP}\label{alg:seqp}
\noindent\makebox[\linewidth]{\rule{\textwidth}{0.4pt}}
{\bf Input}: an approval-based multi-winner election $(N, C, \mathcal{A}, t)$\\
{\bf Output}: the set of winners $W$\\
\noindent\makebox[\linewidth]{\rule{\textwidth}{0.4pt}}

\algblockdefx[ForEach]{ForEach}{EndForEach}[1]
{\textbf{foreach} #1 \textbf{do}}{\textbf{end foreach}} 
{\small
\begin{algorithmic}[1]
\State Compute the set of voters $N_c$ that approve of each candidate $c$, and $|N_c|$
\State $W\gets \emptyset$
\ForEach{$v_i \in N$}
  \State $x_i \gets 0$ \Comment{$x_i$ is the load of voter $v_i$}
\EndForEach

\For{$j=1$ \textbf{to} $t$}
  \ForEach{$c \in C \setminus W$}
    \State $q \gets 0$
    \ForEach{$v_i \in N_c$}
      \State $q \gets q + x_i$
    \EndForEach
    \State $s_c \gets \frac{1+q}{|N_c|}$
  \EndForEach
  \State $w \gets \textrm{argmin} \{s_c: c \in C \setminus W\}$
  \State $W \gets W \cup \{w\}$
  \ForEach{$v_i \in N_w$}
    \State $x_i \gets s_w$
  \EndForEach
\EndFor 
\State \textbf{return} $W$
\end{algorithmic}
}
\end{algorithm}

\section{The voting rules ES and SeqP\label{app:es_seqp}}

In this section, we describe the operation of the method of {\it Equal Shares} (ES)~\citep{peters2020proportionality} and {\it seq-Phragm\'en} (SeqP)~\citep{phragmen:p1},  include pseudocode for computing them, and derive upper bounds on the worst-case time and memory complexity for both rules. 

In the case of the rule ES, initially, each voter possesses one vote, and voters can use fractions of their votes to back candidates. A candidate is added to the set of winners at each iteration.
For that to happen,
the voters that approve of it must possess at least $\frac{n}{t}$ votes altogether.
This amount will
be subtracted from the votes of those voters
at the end of the iteration.

Let $\rho_j(v_i)$ be the fraction of vote possessed by voter $v_i$ after $j$ candidates have been added to
the set of winners. Thus, $\rho_0(v_i)= 1$ for each voter $v_i$.
At iteration $j+1$,
ES adds to the set of winners
the candidate $c$ that minimises $\gamma_{j+1}(c)$,
defined as:

\begin{equation}
\gamma_{j+1}(c) = \min_{x} \left\{x: \sum_{i: c \in A_i} \min \{x,\rho_j(i)\} = \frac{n}{t}\right\}    
\end{equation}

The value of $\rho_{j+1}(v_i)$ is set to $\rho_j(v_i) - \min \{\gamma_{j+1}(c),\rho_j(v_i)\}$ for each voter $v_i$ that approves of candidate
$c$, and to $\rho_j(v_i)$ otherwise. The algorithm stops at the iteration in which no candidate has at least $\frac{n}{t}$ votes anymore, and it outputs the candidates that have been added to the set of winners so far. 

To compute ES efficiently, we follow the procedure suggested in~\cite{peters2020proportionality}. At each iteration, we first sort the votes in increasing order of $\rho(v_i)$. We recall that $\rho_j(v_i)$ is the fraction of vote possessed by voter $v_i$ after $j$ candidates have been added to the set of winners. 

In Algorithm~\ref{alg:es}, we remove the $j$, and refer to $\rho(v_i)$ as the fraction of vote possessed by voter $v_i$ in the current iteration. 

At each iteration $j$ we need to compute the value $\gamma_{j+1}(c)$ of each candidate $c$ that has not yet been added to the set of winners. We recall that $\gamma_{j+1}(c)$ is the minimum fraction of vote, such that if each voter $v_i$ who approves of candidate $c$ contributes to the election of $c$ with $\min\{\gamma_{j+1}(c),\rho(v_i)\}$, then, the sum of all contributions equals to $\frac{n}{t}$.

The value of $\gamma_{j+1}(c)$ can be computed by iteratively accumulating the values of $\rho(v_i)$ of the first voters that approve of candidate $c$ in increasing order of $\rho(v_i)$. Let $n_c$ be the number of voters that approve of candidate $c$, $p$ be the sum of the first $x$ values of $\rho(v_i)$ of voters that approve of candidate $c$. We accumulate the values of $\rho(v_i)$ in $p$ until we find a voter $v_i$, such that $\frac{n}{t} - p  \leq \rho(v_i) (n_c -x)$ and set $\gamma_{j+1}(c)$ to $\frac{\frac{n}{t} - p}{n_c - x}$.      

Algorithm~\ref{alg:es} presents a pseudocode for computing the winners of an election with ES using these ideas. The computation of $\gamma_{j+1}(c)$ for each candidate $c$ is done in the loop in lines $10$ to $26$. Algorithm~\ref{alg:es} stores in variable $\gamma$ the smallest value of $\gamma_{j+1}(c)$ found so far and in variable $\gamma'$ the value of $\gamma_{j+1}(c)$ of the candidate $c$ considered in each iteration of the loop.

By examining Algorithm~\ref{alg:es} we can derive an upper bound for the worst-case time complexity of ES. 
We assume that we have a table of $n \times m$ bits (where $n$ is the number of voters and $m$ is the number of candidates), where in row $i$ the bits of the columns corresponding to candidates approved of by voter $v_i$ are set to $1$ and the others are set to $0$. We also assume that arithmetic operations can be done in constant time. 

Then, the loop in lines $4$ to $6$ can be computed in $O(n)$. The big repeat loop between lines $7$ and $36$ is executed at most $t$ times. In each execution of this loop, we have to sort $n$ reals in line $9$. This can be done in $\mathcal{O}(n \log n)$. Further, the loop from lines $10$ to $26$ must be executed at most $m$ times, and the inner loop from lines $13$ to $25$ must be executed at most $n$ times. Finally, the loop in lines $29$ and $34$ is executed $n$ times. Combining everything, we get that the worst-case complexity of executing ES is bounded by $\mathcal{O}(nt (m+\log n))$. To execute this algorithm, we need to store in memory the table of $n \times m$ bits mentioned above plus the list $\ell$ that contains $n$ pairs $(\rho(v_i),i)$. Thus, the memory complexity of this algorithm is $\mathcal{O}(nm)$. 

SeqP is based on the notion of {\it load}: each candidate added to the set of 
winners induces a unit of load that must be supported among the voters that approve of it. At each iteration,
the candidate that minimises the maximum voter load is chosen. Let $x^j_i$ be the load of voter $v_i$ after the first $j$ candidates have been added to the set of winners ($x^0_i= 0$ for each voter $v_i$).
At this point,
the load $s^{j+1}_c$ each voter $v_i$ that approves of
candidate $c$ would support
in case $c$ were added to the set of winners is:

\begin{equation}
s^{j+1}_c = \frac{1 + \sum_{v_i: c \in A_i} x^j_i}{|\{v_i: c \in A_i\}|} 
\label{eq:seqp}
\end{equation}

The candidate $c$ with the lower $s^{j+1}_c$ is added to the set of winners. Then, voter loads
are updated as follows: $x^{j+1}_i= s^{j+1}_c$ for each voter $v_i$ that approves of $c$, and $x^{j+1}_i= x^{j}_i$
otherwise. The algorithm continues until the set of winners contains exactly $t$ candidates. Therefore, SeqP is not a UTCS rule. \citet{brill2024phragmen} proved that SeqP satisfies PJR but fails EJR. 

In the neural networks experiment, we use SeqP to complete the set of winners output by SEJR or ES (with or without target committee size optimization). To that end, we first need to update the loads of the voters who approve of the winners output by SEJR or ES. In detail, suppose that the outcome of SEJR or ES (with or without target committee size optimization) is $W=\{w_1, \dots,w_{t'}\}$, the desired number of instances per class is $t$, and $t' < t$. To update the loads of the voters that approve of the candidates in W, we iteratively apply Equation~(\ref{eq:seqp}) to compute $s^{j+1}_{w_j}$, and update the load of the voters that approve of candidate $w_j$ to $s^{j+1}_{w_j}$, for $j=1, \dots, t'$. Then, we run SeqP for $t-t'$ iterations to obtain $t-t'$ additional winners.

The pseudocode for computing SeqP is much simpler and given in Algorithm~\ref{alg:seqp}. We simply have to calculate, at each iteration, the value $s_c^{j+1}$ (as described before) for each candidate not yet elected, and select the candidate with the lowest $s_c^{j+1}$. 

We assume again that that we have a table of $n \times m$ bits (where $n$ is the number of voters and $m$ is the number of candidates), where in row $i$ the bits of the columns corresponding to candidates approved of by voter $v_i$ are set to $1$, and the others are set to $0$, and that arithmetic operations can be done in constant time. In line $1$, we compute the set of voters $N_c$ who approve of each candidate (they can be stored, for instance, in a linked list) and $|N_c|$. This can be done in $\mathcal{O}(nm)$. The loop in lines from $3$ to $5$ can be executed in $\mathcal{O}(n)$. Finally, we have to analyze the complexity of computing the big loop in lines from $6$ to $19$ where the winners are selected. The outer loop is executed $t$ times. The loop in lines from $7$ to $13$ is executed at most $m$ times. The loop in lines from $9$ to $11$, and the loop in lines from $16$ to $18$ are executed at most $n$ times. Combining everything we get that the worst-case complexity of Algorithm~\ref{alg:seqp} is bounded by $\mathcal{O}(nmt)$. The memory required to compute Algorithm~\ref{alg:seqp} is in $\mathcal{O}(nm)$ due to the table of $n \times m$ bits.

We also briefly discuss the SEJR case. A pseudocode for computing SEJR is given in~\cite{SEL17a}. There, it is shown that the worst-case time complexity of computing SEJR using the given pseudocode is $\mathcal{O}(nmt)$. The memory necessary to compute SEJR with the pseudocode given in~\cite{SEL17a} consists in a table of $n \times m$ bits to store the approvals of the voters (as in the other two rules), an array of $n$ counters that stores the value of $|A_i \cap W|$ for each voter $i$, and an array in which we store the EJR+ demand of the candidates. Therefore, the memory complexity of executing SEJR is bounded by $\mathcal{O}(nm)$.

\section{Results for S$2$EJR}
\label{app:s2ejr}

We first prove that S$2$EJR outputs a committee of size less
than or equal to $t$.

\begin{lemma}
  
  For any approval-based multi-winner election $(N, C, \mathcal{A},
  t)$, the rule S$2$EJR always outputs a committee of size less than
  or equal to $t$.

\end{lemma}  

\begin{proof}

  We divide the operation of S$2$EJR into two parts: (i) iterations in
  which the maximum (modified) EJR+ demand $\textrm{d}(c,W)$ is greater than or equal to
  $2$ for some candidate $c$; and (ii) iterations in which the maximum
  (modified) EJR+ demand is equal to $1$.

  At each iteration of part (i), a candidate is added to the set of
  winners such that (a) it is approved of by a set of voters $N^*$ of
  size at least $2 \frac{n}{t}$; and (b) each voter in $N^*$ approves
  of at most one of the candidates that have already been added to the
  set of winners. This implies that each voter can belong to $N^*$ at
  most twice in the iterations corresponding to part (i). Suppose that
  part (i) comprises $j$ iterations. Then, at the end of part (i), the
  total number of voters who approve of at least one of the
  candidates in the set of winners are at least:

  \begin{displaymath}
    \frac{j 2 \frac{n}{t}}{2} = j \frac{n}{t} \leq n
  \end{displaymath}.

  This proves that $j$ is not greater than $t$. Now, at each iteration
  of part (ii), a candidate is added to the set of winners such that
  (a) it is approved of by a set of voters $N^*$ of size at least
  $\frac{n}{t}$; and (b) each voter in $N^*$ does not approve of any of
  the candidates that have already been added to the set of
  winners. This implies that such voters can belong to $N^*$ at most
  once in the iterations corresponding to part (ii). Suppose that part
  (ii) comprises $h$ iterations. Then, the total number of voters that
  belong to $N^*$ in some iteration of part (ii) is, at least:

  \begin{displaymath}
    \frac{h \frac{n}{t}}{1} = h \frac{n}{t} \leq n - j \frac{n}{t} 
  \end{displaymath}.

  This implies that $h+j \leq t$.  
  
\end{proof}

\begin{lemma}
  S$2$EJR satisfies $2$-EJR.
\end{lemma}

\begin{proof}

It is straightforward to prove the S$2$EJR satisfies $2$-EJR. For the sake
of contradiction, suppose that for a certain approval-based
multi-winner election $(N, C, \mathcal{A}, t)$, with $t \geq 2$, the
set of winners $W$ that S$2$EJR outputs does not satisfy $2$-EJR.
This implies that there must exist a set of voters $N^*$ such that
$|N^*| \geq 2 \frac{n}{t}$, $|\bigcap_{v_i \in N^*} A_i| \geq 2$, but
$|A_i \cap W| < 2$ for each voter $v_i$ in $N^*$. Furthermore, a candidate
$c$ must exist such that $c$ belongs to $A_i$ for each voter $v_i$ in
$N^*$, but $c$ does not belong to $W$. According to the modified definition of
EJR+ demand for S$2$EJR, the set $N^*$ fulfils all the
requirements so that the (modified) EJR+ demand of candidate $c$ would be at
least $2$. This is a contradiction because S$2$EJR stops when
all the candidates that have not yet been added to the set of winners
have a (modified) EJR+ demand equal to zero.
  
\end{proof}  

\begin{example}
  Consider the following election $(N,C,\mathcal{A},3)$, where $N=
  \{v_1\}$, $C= \{c_1, c_2, c_3\}$, and $A_1= \{c_1, c_2, c_3\}$. For
  this election, $N$ is a $3$-cohesive group of voters, but S$2$EJR
  outputs a set of winners of size only $2$. This proves that S$2$EJR
  fails EJR.
\end{example}

\begin{table}[t]
    \scriptsize
    \setlength{\tabcolsep}{3pt}     
    \renewcommand{\arraystretch}{0.95}

  \begin{tabular}{M{0.1cm}rlrrrrrrrr}
    & \multicolumn{1}{c}{IPC} & \multicolumn{1}{c}{Classifier} & \multicolumn{1}{c}{SEJR} & \multicolumn{1}{c}{ES} & \multicolumn{1}{c}{seq-Phrag.} & \multicolumn{1}{c}{Herding} & \multicolumn{1}{c}{k-Means} & \multicolumn{1}{c}{k-Center} & \multicolumn{1}{c}{Random} \\
    \hline
    \multirow{8}{*}{\makecell{\rotatebox{90}{CIFAR-100}}} & \multirow[t]{4}{*}{50} & ConvNet & $\textbf{33.84} \pm 0.09$ & $33.67 \pm 0.09$ & $33.62 \pm 0.09$ & $32.74 \pm 0.11$ & $31.10 \pm 0.11$ & $23.83 \pm 0.11$ & $30.08 \pm 0.13$ \\

    &  & MLP & $17.99 \pm 0.08$ & $\textbf{18.02} \pm 0.07$ & $17.89 \pm 0.06$ & $15.80 \pm 0.06$ & $14.67 \pm 0.08$ & $9.83 \pm 0.10$ & $14.53 \pm 0.10$ \\

    & & ResNet18 & $18.93 \pm 0.17$ & $\textbf{19.29} \pm 0.20$ & $19.21 \pm 0.15$ & $14.48 \pm 0.19$ & $14.95 \pm 0.16$ & $10.68 \pm 0.12$ & $14.17 \pm 0.28$ \\

    & & ViT & $21.64 \pm 0.11$ & $\textbf{21.72} \pm 0.09$ & $21.48 \pm 0.12$ & $18.15 \pm 0.11$ & $17.49 \pm 0.11$ & $11.45 \pm 0.13$ & $16.60 \pm 0.14$ \\
    \cline{2-10}
     & \multirow[t]{4}{*}{10} & ConvNet & $19.24 \pm 0.07$ & $19.41 \pm 0.09$ & $ \textbf{19.55} \pm 0.08$ & $17.28 \pm 0.09$ & $15.37 \pm 0.08$ & $8.98 \pm 0.12$ & $14.44 \pm 0.14$ \\

    &  & MLP & $12.39 \pm 0.06$ & $12.35 \pm 0.06$ & $\textbf{12.60} \pm 0.06$ & $11.17 \pm 0.06$ & $8.92 \pm 0.06$ & $5.05 \pm 0.05 $ & $8.99 \pm 0.10$ \\

    & & ResNet18 & $8.39 \pm 0.09$ & $\textbf{8.63} \pm 0.12$ & $8.61 \pm 0.10$ & $5.55 \pm 0.11$ & $5.50 \pm 0.10$ & $3.14 \pm 0.08$ & $5.08 \pm 0.11$ \\

            & & ViT & $\textbf{10.12} \pm 0.31$ & $9.56 \pm 0.36$ & $ 10.05 \pm 0.25$ & $6.83 \pm 0.11$ & $6.85 \pm 0.20$ & $3.89 \pm 0.11$ & $5.97 \pm 0.14$ \\

    \hline

    \multirow{8}{*}{\makecell{\rotatebox{90}{CIFAR-10}}} & \multirow[t]{4}{*}{50} & ConvNet & $\textbf{47.00} \pm 0.14$ & $46.12 \pm 0.13$ & $46.63 \pm 0.11$ & $46.13 \pm 0.14$ & $44.61 \pm 0.09$ & $38.25 \pm 0.3$ & $43.72 \pm 0.22$ \\

    &  & MLP & $\textbf{33.04} \pm 0.08$ & $32.92 \pm 0.24$ & $32.93 \pm 0.07$ & $30.72 \pm 0.10$ & $28.87 \pm 0.08$ & $25.71 \pm 0.26$ & $28.87 \pm 0.08$ \\

    & & ResNet18 & $\textbf{32.45} \pm 0.25$ & $32.41 \pm 0.23$ & $32.43 \pm 0.23$ & $28.56 \pm 0.23$ & $28.69 \pm 0.24$ & $23.86 \pm 0.39$ & $27.98 \pm 0.23$ \\

            & & ViT & $31.72 \pm 0.24$ & $31.72 \pm 0.30$ & $\textbf{32.12} \pm 0.29$ & $25.35 \pm 0.26$ & $25.15 \pm 0.29$ & $21.37 \pm 0.24$ & $25.29 \pm 0.30$ \\
    \cline{2-10}
     & \multirow[t]{4}{*}{10} & ConvNet & $32.82 \pm 0.14$ & $\textbf{33.37} \pm 0.15$ & $32.92 \pm 0.11$ & $26.69 \pm 0.19$ & $31.15 \pm 0.20$ & $25.14 \pm 0.35$ & $26.81 \pm 0.63$ \\

    &  & MLP & $27.03 \pm 0.15$ & $\textbf{27.59} \pm 0.10$ & $26.54 \pm 0.10$ & $22.85 \pm 0.12$ & $25.43 \pm 0.13$ & $20.96 \pm 0.43$ & $22.47 \pm 0.47$ \\

    & & ResNet18 & $25.12 \pm 0.25$ & $25.37 \pm 0.18$ & $\textbf{25.87} \pm 0.20 $ & $20.62 \pm 0.20$ & $23.14 \pm 0.27$ & $18.10 \pm 0.37$ & $20.08 \pm 0.41$ \\

            & & ViT & $ \textbf{24.66} \pm 0.23$ & $23.12 \pm 0.13$ & $22.98 \pm 0.11$ & $21.55 \pm 0.12$ & $21.54 \pm 0.14$ & $18.96 \pm 0.22$ & $20.02 \pm 0.50$ \\
\hline
  \end{tabular}
  \caption{Average accuracy results for experiment I,
    with 95\% confidence intervals. Alternative SEJR and ES classifiers,
  as explained in Section~\ref{app:summary}.}
  \label{tab:exp-1-accuracyV2}
\end{table}

\section{Statistical analysis of Experiment II: omitted details}
\label{sec:suppl_statistic}

The first step in evaluating the algorithms is to compute the value for a given metric (accuracy for KNN or SVM, or reduction) obtained with each algorithm for a certain number of datasets. We rank the algorithms for each dataset according to the values for a given metric (accuracy for KNN or SVM, or reduction) obtained by each algorithm for such a dataset. We assign rank $1$ to the algorithm that achieves the best performance for that dataset, rank $2$ to the second-best, and so on. If several algorithms are tied, we assign each of them the average of all the positions that must be assigned to them. For instance, if for a particular dataset the best performance is obtained with algorithm A, and B, C, and D are tied as the second best algorithms in performance just after A, then algorithm A is assigned rank $1$, and algorithms B, C, and D are all assigned rank $\frac{2+3+4}{3}= 3$.

We have $k$ algorithms and $N$ datasets. Let $r_i^j$ be the ranking of algorithm $j$ for dataset $i$. 
Each algorithm is assigned an average ranking, defined as: 

\begin{equation}
  R_j= \frac{1}{N} \sum_i r_i^j
\end{equation}

Now, the Friedman statistic is defined as:

\begin{equation}
  \chi^2_F \sim \frac{12N}{k(k+1)} \Big[ \Big( \sum_j R_j^2 \Big) -
    \frac{k(k+1)^2}{4} \Big],
\end{equation}

where $\chi^2_F$ follows a Friedman $\chi^2$ distribution with 
parameters $k$ and $N$. For large enough values of
$k$ and $N$, the Friedman $\chi^2$ distribution converges to the
standard $\chi^2$ distribution with $k-1$ degrees of freedom.

To reject the null hypothesis that all the algorithms considered are equivalent for a certain metric
with significance level $\alpha$, we compare the value  $x$ obtained with the Friedman statistic with the value $y$ of
the $\chi^2_F$ distribution with parameters $k$ and $N$ corresponding to a p-value equal to $\alpha$. Then, if $x > y$, we can reject the null hypothesis.

If the Friedman test allows the rejection of the null hypothesis, a post-hoc test can be conducted to compare the performance of each pair of algorithms. Following the methodology recommended by~\citet{garcia2008extension}, for comparing the performance of each pair of algorithms, we use the~\citet{bergmann1988improvements} test. 
It allows us to accept or reject the null hypothesis that two algorithms are equivalent according to a given metric (accuracy for KNN or SVM, or reduction). If the null hypothesis is rejected, then the algorithm with a better average ranking (over all the datasets) is considered {\it statistically better} than the other algorithm for the metric (accuracy for KNN or SVM, or reduction) considered.

Bergmann and Hommel's test uses the following statistic to compare algorithms:

\begin{equation}
  z \sim \frac{R_i - R_j}{\sqrt{\frac{k(k+1)}{6N}}}
  \label{est:Holm}
\end{equation}

Where $z$ follows a normal distribution $N(0,1)$. The test of Bergmann and Hommel is based on the notion of {\it exhaustive sets}. The idea of exhaustive sets is that not all sets of hypotheses are feasible. Suppose, for instance, that we have three algorithms $A$, $B$ and $C$, and we want to know which algorithms are equivalent for a certain metric. Suppose we also know that algorithm $A$ is better than algorithm $B$. Then, it is not possible that both `$A$ is equivalent to $C$' and `$B$ is equivalent to $C$' hold. In other words, $\{AC, BC\}$ is not a set of feasible hypotheses (pairs of letters mean pairs of algorithms that are equivalent). 
In contrast, both $\{AB, AC, BC\}$ and $\{BC\}$ are feasible sets of hypotheses. Feasible sets of hypotheses are called exhaustive sets by Bergmann and Hommel. For details about the computation of exhaustive sets, we refer to~\citet{garcia2008extension,bergmann1988improvements}.

For each exhaustive set $I$, the test of Bergmann and Hommel accepts all the hypotheses contained in $I$ if $\min \{p_i: i \in I\} > \frac{\alpha}{|I|}$, where $p_i$ is the p-value of hypotheses $i$, obtained from the statistic~(\ref{est:Holm}), and $\alpha$ is the significance level. 

\subsection{Results}

The computation of the exhaustive sets grows exponentially with the number of algorithms to compare~\citep{garcia2008extension}. We have only been able to compute the exhaustive sets for up to $13$ algorithms. For $13$ algorithms, we have $27,644,436$ different exhaustive sets that, when stored in main memory, consume about 48 gigabytes. Table~4 in the main paper considers $11$ baselines and state-of-the-art instance selection algorithms. 
Therefore, we decided to perform the statistical analysis 12 times. Each statistical analysis considers the $11$ baselines and state-of-the-art instance selection algorithms considered in Table~4 in the main paper, plus the algorithms included in Table~3 in the main paper for each of the four rules considered: SEJR, ES, SeqP, and S$2$EJR. For each of these $13$ combinations of algorithms, we performed the statistical analysis for accuracy with SVM, accuracy with KNN, and reduction.

We use in all cases a significance level $\alpha=0.05$.
Since all the statistical analysis compare $13$ algorithms using $15$ datasets, the value $y$ of
the $\chi^2_F$ distribution with parameters $k= 13$ and $N= 15$ corresponding to a p-value equal to $\alpha= 0.05$ is the same for the $12$ statistical analyses considered. Accoding to~\citet{lopez2019extended}, the $\chi^2_F$ distribution with parameters $k= 13$ and $N= 15$ can be approximated with a standard $\chi^2$ distribution with $12$ degrees of freedom with an error of less than $10\%$. Using the standard $\chi^2$ distribution with $12$ degrees of freedom, we obtain an approximation of $y \sim 21.026$.  

\begin{table}[!thb]
\begin{center}
\setlength{\tabcolsep}{2pt}
\begin{tabular}{lrrrr}
\toprule
metric&SEJR&ES&SeqP&S$2$EJR \\
\midrule
accuracy (SVM) &  119.516 & 113.292 & 118.220 & 107.899 \\
accuracy (KNN) & 85.681 & 80.952 & 84.288 & 82.497 \\
reduction & 158.738 & 155.424 & 118.220 & 157.804 \\
\bottomrule
\end{tabular}
\end{center}
\caption{Value of the Friedman statistic obtained in each statistical analysis}
\label{t:FriedmanStValues}
\end{table}

Table~\ref{t:FriedmanStValues} shows, for each statistical analysis, the value of the Friedman statistic. Labels in the columns indicate the rule that is included in the algorithms considered in the statistical analysis, and rows indicate the metric considered in the statistical analysis. 
In summary, the null hypothesis that all the algorithms are equivalent can be discarded with a significance level $\alpha= 0.05$ in all the statistical analyses.

Therefore, in all cases, we can conduct a post-hoc test to compare the performance of each pair of algorithms using the~\citet{bergmann1988improvements} test. The results of such tests are shown in Tables~\ref{t:statResultsSEJR-SVM} to~\ref{t:statResultsSeqP-red}.

\begin{sidewaystable}
\begin{center}
\setlength{\tabcolsep}{2pt}
\begin{tabular}{llllllllllllll}
\toprule
Algorithm & NoR & NoA & R-0.9 & R-0.1 & SEJR-2 & SEJR-0.25 & DROP3 & ENN & ICF & LSBo & LSSm & LDIS & ISDSP \\
\midrule
NoR & $\sim$ & $\sim$ & $\sim$ & $>$ & $\sim$ & $>$ & $>$ & $\sim$ & $>$ & $>$ & $\sim$ & $>$ & $>$ \\
NoA & $\sim$ & $\sim$ & $\sim$ & $>$ & $\sim$ & $>$ & $\sim$ & $\sim$ & $\sim$ & $>$ & $\sim$ & $\sim$ & $>$ \\
R-0.9 & $\sim$ & $\sim$ & $\sim$ & $>$ & $\sim$ & $>$ & $\sim$ & $\sim$ & $>$ & $>$ & $\sim$ & $>$ & $>$ \\
R-0.1 & $<$ & $<$ & $<$ & $\sim$ & $<$ & $\sim$ & $\sim$ & $<$ & $\sim$ & $\sim$ & $<$ & $\sim$ & $\sim$ \\
SEJR-2 & $\sim$ & $\sim$ & $\sim$ & $>$ & $\sim$ & $>$ & $>$ & $\sim$ & $>$ & $>$ & $\sim$ & $>$ & $>$ \\
SEJR-0.25 & $<$ & $<$ & $<$ & $\sim$ & $<$ & $\sim$ & $\sim$ & $<$ & $\sim$ & $\sim$ & $<$ & $\sim$ & $\sim$ \\
DROP3 & $<$ & $\sim$ & $\sim$ & $\sim$ & $<$ & $\sim$ & $\sim$ & $\sim$ & $\sim$ & $\sim$ & $<$ & $\sim$ & $\sim$ \\
ENN & $\sim$ & $\sim$ & $\sim$ & $>$ & $\sim$ & $>$ & $\sim$ & $\sim$ & $\sim$ & $>$ & $\sim$ & $>$ & $>$ \\
ICF & $<$ & $\sim$ & $<$ & $\sim$ & $<$ & $\sim$ & $\sim$ & $\sim$ & $\sim$ & $\sim$ & $<$ & $\sim$ & $\sim$ \\
LSBo & $<$ & $<$ & $<$ & $\sim$ & $<$ & $\sim$ & $\sim$ & $<$ & $\sim$ & $\sim$ & $<$ & $\sim$ & $\sim$ \\
LSSm & $\sim$ & $\sim$ & $\sim$ & $>$ & $\sim$ & $>$ & $>$ & $\sim$ & $>$ & $>$ & $\sim$ & $>$ & $>$ \\
LDIS & $<$ & $\sim$ & $<$ & $\sim$ & $<$ & $\sim$ & $\sim$ & $<$ & $\sim$ & $\sim$ & $<$ & $\sim$ & $\sim$ \\
ISDSP & $<$ & $<$ & $<$ & $\sim$ & $<$ & $\sim$ & $\sim$ & $<$ & $\sim$ & $\sim$ & $<$ & $\sim$ & $\sim$ \\
\bottomrule
\end{tabular}
\end{center}
\caption{Statistical analysis results with SEJR for accuracy with an SVM classifier}
\label{t:statResultsSEJR-SVM}
\end{sidewaystable}

\begin{sidewaystable}
\begin{center}
\setlength{\tabcolsep}{2pt}
\begin{tabular}{llllllllllllll}
\toprule
Algorithm & NoR & NoA & R-0.9 & R-0.1 & SEJR-2 & SEJR-0.25 & DROP3 & ENN & ICF & LSBo & LSSm & LDIS & ISDSP \\
\midrule
NoR & $\sim$ & $\sim$ & $\sim$ & $>$ & $\sim$ & $>$ & $\sim$ & $\sim$ & $>$ & $\sim$ & $\sim$ & $>$ & $>$ \\
NoA & $\sim$ & $\sim$ & $\sim$ & $>$ & $\sim$ & $>$ & $\sim$ & $\sim$ & $>$ & $\sim$ & $\sim$ & $\sim$ & $>$ \\
R-0.9 & $\sim$ & $\sim$ & $\sim$ & $>$ & $\sim$ & $>$ & $\sim$ & $\sim$ & $>$ & $\sim$ & $\sim$ & $>$ & $>$ \\
R-0.1 & $<$ & $<$ & $<$ & $\sim$ & $<$ & $\sim$ & $<$ & $<$ & $\sim$ & $\sim$ & $<$ & $\sim$ & $\sim$ \\
SEJR-2 & $\sim$ & $\sim$ & $\sim$ & $>$ & $\sim$ & $>$ & $\sim$ & $\sim$ & $>$ & $\sim$ & $\sim$ & $\sim$ & $>$ \\
SEJR-0.25 & $<$ & $<$ & $<$ & $\sim$ & $<$ & $\sim$ & $\sim$ & $\sim$ & $\sim$ & $\sim$ & $<$ & $\sim$ & $\sim$ \\
DROP3 & $\sim$ & $\sim$ & $\sim$ & $>$ & $\sim$ & $\sim$ & $\sim$ & $\sim$ & $\sim$ & $\sim$ & $\sim$ & $\sim$ & $\sim$ \\
ENN & $\sim$ & $\sim$ & $\sim$ & $>$ & $\sim$ & $\sim$ & $\sim$ & $\sim$ & $\sim$ & $\sim$ & $\sim$ & $\sim$ & $\sim$ \\
ICF & $<$ & $<$ & $<$ & $\sim$ & $<$ & $\sim$ & $\sim$ & $\sim$ & $\sim$ & $\sim$ & $<$ & $\sim$ & $\sim$ \\
LSBo & $\sim$ & $\sim$ & $\sim$ & $\sim$ & $\sim$ & $\sim$ & $\sim$ & $\sim$ & $\sim$ & $\sim$ & $\sim$ & $\sim$ & $\sim$ \\
LSSm & $\sim$ & $\sim$ & $\sim$ & $>$ & $\sim$ & $>$ & $\sim$ & $\sim$ & $>$ & $\sim$ & $\sim$ & $>$ & $>$ \\
LDIS & $<$ & $\sim$ & $<$ & $\sim$ & $\sim$ & $\sim$ & $\sim$ & $\sim$ & $\sim$ & $\sim$ & $<$ & $\sim$ & $\sim$ \\
ISDSP & $<$ & $<$ & $<$ & $\sim$ & $<$ & $\sim$ & $\sim$ & $\sim$ & $\sim$ & $\sim$ & $<$ & $\sim$ & $\sim$ \\
\bottomrule
\end{tabular}
\end{center}
\caption{Statistical analysis results with SEJR for accuracy with a KNN classifier}
\label{t:statResultsSEJR-KNN}
\end{sidewaystable}

\begin{sidewaystable}
\begin{center}
\setlength{\tabcolsep}{2pt}
\begin{tabular}{llllllllllllll}
\toprule
Algorithm & NoR & NoA & R-0.9 & R-0.1 & SEJR-2 & SEJR-0.25 & DROP3 & ENN & ICF & LSBo & LSSm & LDIS & ISDSP \\
\midrule
NoR & $\sim$ & $\sim$ & $\sim$ & $<$ & $<$ & $<$ & $<$ & $\sim$ & $<$ & $<$ & $\sim$ & $<$ & $<$ \\
NoA & $\sim$ & $\sim$ & $\sim$ & $<$ & $\sim$ & $<$ & $\sim$ & $\sim$ & $<$ & $<$ & $\sim$ & $<$ & $<$ \\
R-0.9 & $\sim$ & $\sim$ & $\sim$ & $<$ & $\sim$ & $<$ & $\sim$ & $\sim$ & $<$ & $<$ & $\sim$ & $<$ & $<$ \\
R-0.1 & $>$ & $>$ & $>$ & $\sim$ & $>$ & $\sim$ & $\sim$ & $>$ & $\sim$ & $\sim$ & $>$ & $\sim$ & $\sim$ \\
SEJR-2 & $>$ & $\sim$ & $\sim$ & $<$ & $\sim$ & $<$ & $\sim$ & $\sim$ & $\sim$ & $\sim$ & $\sim$ & $<$ & $<$ \\
SEJR-0.25 & $>$ & $>$ & $>$ & $\sim$ & $>$ & $\sim$ & $\sim$ & $>$ & $\sim$ & $\sim$ & $>$ & $\sim$ & $\sim$ \\
DROP3 & $>$ & $\sim$ & $\sim$ & $\sim$ & $\sim$ & $\sim$ & $\sim$ & $\sim$ & $\sim$ & $\sim$ & $>$ & $\sim$ & $\sim$ \\
ENN & $\sim$ & $\sim$ & $\sim$ & $<$ & $\sim$ & $<$ & $\sim$ & $\sim$ & $<$ & $<$ & $\sim$ & $<$ & $<$ \\
ICF & $>$ & $>$ & $>$ & $\sim$ & $\sim$ & $\sim$ & $\sim$ & $>$ & $\sim$ & $\sim$ & $>$ & $\sim$ & $\sim$ \\
LSBo & $>$ & $>$ & $>$ & $\sim$ & $\sim$ & $\sim$ & $\sim$ & $>$ & $\sim$ & $\sim$ & $>$ & $\sim$ & $\sim$ \\
LSSm & $\sim$ & $\sim$ & $\sim$ & $<$ & $\sim$ & $<$ & $<$ & $\sim$ & $<$ & $<$ & $\sim$ & $<$ & $<$ \\
LDIS & $>$ & $>$ & $>$ & $\sim$ & $>$ & $\sim$ & $\sim$ & $>$ & $\sim$ & $\sim$ & $>$ & $\sim$ & $\sim$ \\
ISDSP & $>$ & $>$ & $>$ & $\sim$ & $>$ & $\sim$ & $\sim$ & $>$ & $\sim$ & $\sim$ & $>$ & $\sim$ & $\sim$ \\
\bottomrule
\end{tabular}
\end{center}
\caption{Statistical analysis results with SEJR for reduction}
\label{t:statResultsSEJR-red}
\end{sidewaystable}

\begin{sidewaystable}
\begin{center}
\setlength{\tabcolsep}{2pt}
\begin{tabular}{llllllllllllll}
\toprule
Algorithm & NoR & NoA & R-0.9 & R-0.1 & S2EJR-2 & S2EJR-0.5 & DROP3 & ENN & ICF & LSBo & LSSm & LDIS & ISDSP \\
\midrule
NoR & $\sim$ & $\sim$ & $\sim$ & $>$ & $\sim$ & $>$ & $>$ & $\sim$ & $>$ & $>$ & $\sim$ & $>$ & $>$ \\
NoA & $\sim$ & $\sim$ & $\sim$ & $>$ & $\sim$ & $\sim$ & $\sim$ & $\sim$ & $\sim$ & $>$ & $\sim$ & $>$ & $>$ \\
R-0.9 & $\sim$ & $\sim$ & $\sim$ & $>$ & $\sim$ & $>$ & $\sim$ & $\sim$ & $>$ & $>$ & $\sim$ & $>$ & $>$ \\
R-0.1 & $<$ & $<$ & $<$ & $\sim$ & $<$ & $\sim$ & $\sim$ & $<$ & $\sim$ & $\sim$ & $<$ & $\sim$ & $\sim$ \\
S2EJR-2 & $\sim$ & $\sim$ & $\sim$ & $>$ & $\sim$ & $\sim$ & $\sim$ & $\sim$ & $\sim$ & $\sim$ & $\sim$ & $\sim$ & $\sim$ \\
S2EJR-0.5 & $<$ & $\sim$ & $<$ & $\sim$ & $\sim$ & $\sim$ & $\sim$ & $\sim$ & $\sim$ & $\sim$ & $<$ & $\sim$ & $\sim$ \\
DROP3 & $<$ & $\sim$ & $\sim$ & $\sim$ & $\sim$ & $\sim$ & $\sim$ & $\sim$ & $\sim$ & $\sim$ & $<$ & $\sim$ & $\sim$ \\
ENN & $\sim$ & $\sim$ & $\sim$ & $>$ & $\sim$ & $\sim$ & $\sim$ & $\sim$ & $\sim$ & $>$ & $\sim$ & $>$ & $>$ \\
ICF & $<$ & $\sim$ & $<$ & $\sim$ & $\sim$ & $\sim$ & $\sim$ & $\sim$ & $\sim$ & $\sim$ & $<$ & $\sim$ & $\sim$ \\
LSBo & $<$ & $<$ & $<$ & $\sim$ & $\sim$ & $\sim$ & $\sim$ & $<$ & $\sim$ & $\sim$ & $<$ & $\sim$ & $\sim$ \\
LSSm & $\sim$ & $\sim$ & $\sim$ & $>$ & $\sim$ & $>$ & $>$ & $\sim$ & $>$ & $>$ & $\sim$ & $>$ & $>$ \\
LDIS & $<$ & $<$ & $<$ & $\sim$ & $\sim$ & $\sim$ & $\sim$ & $<$ & $\sim$ & $\sim$ & $<$ & $\sim$ & $\sim$ \\
ISDSP & $<$ & $<$ & $<$ & $\sim$ & $\sim$ & $\sim$ & $\sim$ & $<$ & $\sim$ & $\sim$ & $<$ & $\sim$ & $\sim$ \\
\bottomrule
\end{tabular}
\end{center}
\caption{Statistical analysis results with S2EJR for accuracy with an SVM classifier}
\label{t:statResultsS2EJR-SVM}
\end{sidewaystable}

\begin{sidewaystable}
\begin{center}
\setlength{\tabcolsep}{2pt}
\begin{tabular}{llllllllllllll}
\toprule
Algorithm & NoR & NoA & R-0.9 & R-0.1 & S2EJR-2 & S2EJR-0.5 & DROP3 & ENN & ICF & LSBo & LSSm & LDIS & ISDSP \\
\midrule
NoR & $\sim$ & $\sim$ & $\sim$ & $>$ & $\sim$ & $\sim$ & $\sim$ & $\sim$ & $>$ & $\sim$ & $\sim$ & $>$ & $>$ \\
NoA & $\sim$ & $\sim$ & $\sim$ & $>$ & $\sim$ & $\sim$ & $\sim$ & $\sim$ & $>$ & $\sim$ & $\sim$ & $>$ & $>$ \\
R-0.9 & $\sim$ & $\sim$ & $\sim$ & $>$ & $\sim$ & $\sim$ & $\sim$ & $\sim$ & $>$ & $\sim$ & $\sim$ & $>$ & $>$ \\
R-0.1 & $<$ & $<$ & $<$ & $\sim$ & $<$ & $<$ & $<$ & $<$ & $\sim$ & $\sim$ & $<$ & $\sim$ & $\sim$ \\
S2EJR-2 & $\sim$ & $\sim$ & $\sim$ & $>$ & $\sim$ & $\sim$ & $\sim$ & $\sim$ & $>$ & $\sim$ & $\sim$ & $\sim$ & $>$ \\
S2EJR-0.5 & $\sim$ & $\sim$ & $\sim$ & $>$ & $\sim$ & $\sim$ & $\sim$ & $\sim$ & $\sim$ & $\sim$ & $\sim$ & $\sim$ & $\sim$ \\
DROP3 & $\sim$ & $\sim$ & $\sim$ & $>$ & $\sim$ & $\sim$ & $\sim$ & $\sim$ & $\sim$ & $\sim$ & $\sim$ & $\sim$ & $\sim$ \\
ENN & $\sim$ & $\sim$ & $\sim$ & $>$ & $\sim$ & $\sim$ & $\sim$ & $\sim$ & $\sim$ & $\sim$ & $\sim$ & $\sim$ & $\sim$ \\
ICF & $<$ & $<$ & $<$ & $\sim$ & $<$ & $\sim$ & $\sim$ & $\sim$ & $\sim$ & $\sim$ & $<$ & $\sim$ & $\sim$ \\
LSBo & $\sim$ & $\sim$ & $\sim$ & $\sim$ & $\sim$ & $\sim$ & $\sim$ & $\sim$ & $\sim$ & $\sim$ & $\sim$ & $\sim$ & $\sim$ \\
LSSm & $\sim$ & $\sim$ & $\sim$ & $>$ & $\sim$ & $\sim$ & $\sim$ & $\sim$ & $>$ & $\sim$ & $\sim$ & $>$ & $>$ \\
LDIS & $<$ & $<$ & $<$ & $\sim$ & $\sim$ & $\sim$ & $\sim$ & $\sim$ & $\sim$ & $\sim$ & $<$ & $\sim$ & $\sim$ \\
ISDSP & $<$ & $<$ & $<$ & $\sim$ & $<$ & $\sim$ & $\sim$ & $\sim$ & $\sim$ & $\sim$ & $<$ & $\sim$ & $\sim$ \\
\bottomrule
\end{tabular}
\end{center}
\caption{Statistical analysis results with S2EJR for accuracy with a KNN classifier}
\label{t:statResultsS2EJR-KNN}
\end{sidewaystable}

\begin{sidewaystable}
\begin{center}
\setlength{\tabcolsep}{2pt}
\begin{tabular}{llllllllllllll}
\toprule
Algorithm & NoR & NoA & R-0.9 & R-0.1 & S2EJR-2 & S2EJR-0.5 & DROP3 & ENN & ICF & LSBo & LSSm & LDIS & ISDSP \\
\midrule
NoR & $\sim$ & $\sim$ & $\sim$ & $<$ & $<$ & $<$ & $<$ & $\sim$ & $<$ & $<$ & $\sim$ & $<$ & $<$ \\
NoA & $\sim$ & $\sim$ & $\sim$ & $<$ & $\sim$ & $<$ & $\sim$ & $\sim$ & $<$ & $<$ & $\sim$ & $<$ & $<$ \\
R-0.9 & $\sim$ & $\sim$ & $\sim$ & $<$ & $\sim$ & $<$ & $\sim$ & $\sim$ & $<$ & $<$ & $\sim$ & $<$ & $<$ \\
R-0.1 & $>$ & $>$ & $>$ & $\sim$ & $\sim$ & $\sim$ & $\sim$ & $>$ & $\sim$ & $\sim$ & $>$ & $\sim$ & $\sim$ \\
S2EJR-2 & $>$ & $\sim$ & $\sim$ & $\sim$ & $\sim$ & $<$ & $\sim$ & $\sim$ & $\sim$ & $\sim$ & $\sim$ & $\sim$ & $\sim$ \\
S2EJR-0.5 & $>$ & $>$ & $>$ & $\sim$ & $>$ & $\sim$ & $\sim$ & $>$ & $\sim$ & $\sim$ & $>$ & $\sim$ & $\sim$ \\
DROP3 & $>$ & $\sim$ & $\sim$ & $\sim$ & $\sim$ & $\sim$ & $\sim$ & $\sim$ & $\sim$ & $\sim$ & $\sim$ & $\sim$ & $\sim$ \\
ENN & $\sim$ & $\sim$ & $\sim$ & $<$ & $\sim$ & $<$ & $\sim$ & $\sim$ & $<$ & $<$ & $\sim$ & $<$ & $<$ \\
ICF & $>$ & $>$ & $>$ & $\sim$ & $\sim$ & $\sim$ & $\sim$ & $>$ & $\sim$ & $\sim$ & $>$ & $\sim$ & $\sim$ \\
LSBo & $>$ & $>$ & $>$ & $\sim$ & $\sim$ & $\sim$ & $\sim$ & $>$ & $\sim$ & $\sim$ & $>$ & $\sim$ & $\sim$ \\
LSSm & $\sim$ & $\sim$ & $\sim$ & $<$ & $\sim$ & $<$ & $\sim$ & $\sim$ & $<$ & $<$ & $\sim$ & $<$ & $<$ \\
LDIS & $>$ & $>$ & $>$ & $\sim$ & $\sim$ & $\sim$ & $\sim$ & $>$ & $\sim$ & $\sim$ & $>$ & $\sim$ & $\sim$ \\
ISDSP & $>$ & $>$ & $>$ & $\sim$ & $\sim$ & $\sim$ & $\sim$ & $>$ & $\sim$ & $\sim$ & $>$ & $\sim$ & $\sim$ \\
\bottomrule
\end{tabular}
\end{center}
\caption{Statistical analysis results with S2EJR for reduction}
\label{t:statResultsS2EJR-red}
\end{sidewaystable}

\begin{sidewaystable}
\begin{center}
\setlength{\tabcolsep}{2pt}
\begin{tabular}{llllllllllllll}
\toprule
Algorithm & NoR & NoA & R-0.9 & R-0.1 & ES-2 & ES-0.25 & DROP3 & ENN & ICF & LSBo & LSSm & LDIS & ISDSP \\
\midrule
NoR & $\sim$ & $\sim$ & $\sim$ & $>$ & $\sim$ & $>$ & $>$ & $\sim$ & $>$ & $>$ & $\sim$ & $>$ & $>$ \\
NoA & $\sim$ & $\sim$ & $\sim$ & $>$ & $\sim$ & $\sim$ & $\sim$ & $\sim$ & $\sim$ & $>$ & $\sim$ & $\sim$ & $>$ \\
R-0.9 & $\sim$ & $\sim$ & $\sim$ & $>$ & $\sim$ & $>$ & $\sim$ & $\sim$ & $>$ & $>$ & $\sim$ & $>$ & $>$ \\
R-0.1 & $<$ & $<$ & $<$ & $\sim$ & $<$ & $\sim$ & $\sim$ & $<$ & $\sim$ & $\sim$ & $<$ & $\sim$ & $\sim$ \\
ES-2 & $\sim$ & $\sim$ & $\sim$ & $>$ & $\sim$ & $>$ & $\sim$ & $\sim$ & $\sim$ & $>$ & $\sim$ & $>$ & $>$ \\
ES-0.25 & $<$ & $\sim$ & $<$ & $\sim$ & $<$ & $\sim$ & $\sim$ & $<$ & $\sim$ & $\sim$ & $<$ & $\sim$ & $\sim$ \\
DROP3 & $<$ & $\sim$ & $\sim$ & $\sim$ & $\sim$ & $\sim$ & $\sim$ & $\sim$ & $\sim$ & $\sim$ & $<$ & $\sim$ & $\sim$ \\
ENN & $\sim$ & $\sim$ & $\sim$ & $>$ & $\sim$ & $>$ & $\sim$ & $\sim$ & $\sim$ & $>$ & $\sim$ & $>$ & $>$ \\
ICF & $<$ & $\sim$ & $<$ & $\sim$ & $\sim$ & $\sim$ & $\sim$ & $\sim$ & $\sim$ & $\sim$ & $<$ & $\sim$ & $\sim$ \\
LSBo & $<$ & $<$ & $<$ & $\sim$ & $<$ & $\sim$ & $\sim$ & $<$ & $\sim$ & $\sim$ & $<$ & $\sim$ & $\sim$ \\
LSSm & $\sim$ & $\sim$ & $\sim$ & $>$ & $\sim$ & $>$ & $>$ & $\sim$ & $>$ & $>$ & $\sim$ & $>$ & $>$ \\
LDIS & $<$ & $\sim$ & $<$ & $\sim$ & $<$ & $\sim$ & $\sim$ & $<$ & $\sim$ & $\sim$ & $<$ & $\sim$ & $\sim$ \\
ISDSP & $<$ & $<$ & $<$ & $\sim$ & $<$ & $\sim$ & $\sim$ & $<$ & $\sim$ & $\sim$ & $<$ & $\sim$ & $\sim$ \\
\bottomrule
\end{tabular}
\end{center}
\caption{Statistical analysis results with ES for accuracy with an SVM classifier}
\label{t:statResultsES-SVM}
\end{sidewaystable}

\begin{sidewaystable}
\begin{center}
\setlength{\tabcolsep}{2pt}
\begin{tabular}{llllllllllllll}
\toprule
Algorithm & NoR & NoA & R-0.9 & R-0.1 & ES-2 & ES-0.25 & DROP3 & ENN & ICF & LSBo & LSSm & LDIS & ISDSP \\
\midrule
NoR & $\sim$ & $\sim$ & $\sim$ & $>$ & $\sim$ & $>$ & $\sim$ & $\sim$ & $>$ & $\sim$ & $\sim$ & $>$ & $>$ \\
NoA & $\sim$ & $\sim$ & $\sim$ & $>$ & $\sim$ & $>$ & $\sim$ & $\sim$ & $>$ & $\sim$ & $\sim$ & $\sim$ & $>$ \\
R-0.9 & $\sim$ & $\sim$ & $\sim$ & $>$ & $\sim$ & $>$ & $\sim$ & $\sim$ & $>$ & $\sim$ & $\sim$ & $>$ & $>$ \\
R-0.1 & $<$ & $<$ & $<$ & $\sim$ & $<$ & $\sim$ & $<$ & $<$ & $\sim$ & $\sim$ & $<$ & $\sim$ & $\sim$ \\
ES-2 & $\sim$ & $\sim$ & $\sim$ & $>$ & $\sim$ & $\sim$ & $\sim$ & $\sim$ & $>$ & $\sim$ & $\sim$ & $\sim$ & $>$ \\
ES-0.25 & $<$ & $<$ & $<$ & $\sim$ & $\sim$ & $\sim$ & $\sim$ & $\sim$ & $\sim$ & $\sim$ & $<$ & $\sim$ & $\sim$ \\
DROP3 & $\sim$ & $\sim$ & $\sim$ & $>$ & $\sim$ & $\sim$ & $\sim$ & $\sim$ & $\sim$ & $\sim$ & $\sim$ & $\sim$ & $\sim$ \\
ENN & $\sim$ & $\sim$ & $\sim$ & $>$ & $\sim$ & $\sim$ & $\sim$ & $\sim$ & $\sim$ & $\sim$ & $\sim$ & $\sim$ & $\sim$ \\
ICF & $<$ & $<$ & $<$ & $\sim$ & $<$ & $\sim$ & $\sim$ & $\sim$ & $\sim$ & $\sim$ & $<$ & $\sim$ & $\sim$ \\
LSBo & $\sim$ & $\sim$ & $\sim$ & $\sim$ & $\sim$ & $\sim$ & $\sim$ & $\sim$ & $\sim$ & $\sim$ & $\sim$ & $\sim$ & $\sim$ \\
LSSm & $\sim$ & $\sim$ & $\sim$ & $>$ & $\sim$ & $>$ & $\sim$ & $\sim$ & $>$ & $\sim$ & $\sim$ & $>$ & $>$ \\
LDIS & $<$ & $\sim$ & $<$ & $\sim$ & $\sim$ & $\sim$ & $\sim$ & $\sim$ & $\sim$ & $\sim$ & $<$ & $\sim$ & $\sim$ \\
ISDSP & $<$ & $<$ & $<$ & $\sim$ & $<$ & $\sim$ & $\sim$ & $\sim$ & $\sim$ & $\sim$ & $<$ & $\sim$ & $\sim$ \\
\bottomrule
\end{tabular}
\end{center}
\caption{Statistical analysis results with ES for accuracy with a KNN classifier}
\label{t:statResultsES-KNN}
\end{sidewaystable}

\begin{sidewaystable}
\begin{center}
\setlength{\tabcolsep}{2pt}
\begin{tabular}{llllllllllllll}
\toprule
Algorithm & NoR & NoA & R-0.9 & R-0.1 & ES-2 & ES-0.25 & DROP3 & ENN & ICF & LSBo & LSSm & LDIS & ISDSP \\
\midrule
NoR & $\sim$ & $\sim$ & $\sim$ & $<$ & $\sim$ & $<$ & $<$ & $\sim$ & $<$ & $<$ & $\sim$ & $<$ & $<$ \\
NoA & $\sim$ & $\sim$ & $\sim$ & $<$ & $\sim$ & $<$ & $\sim$ & $\sim$ & $<$ & $<$ & $\sim$ & $<$ & $<$ \\
R-0.9 & $\sim$ & $\sim$ & $\sim$ & $<$ & $\sim$ & $<$ & $\sim$ & $\sim$ & $<$ & $<$ & $\sim$ & $<$ & $<$ \\
R-0.1 & $>$ & $>$ & $>$ & $\sim$ & $>$ & $\sim$ & $\sim$ & $>$ & $\sim$ & $\sim$ & $>$ & $\sim$ & $\sim$ \\
ES-2 & $\sim$ & $\sim$ & $\sim$ & $<$ & $\sim$ & $<$ & $\sim$ & $\sim$ & $<$ & $<$ & $\sim$ & $<$ & $<$ \\
ES-0.25 & $>$ & $>$ & $>$ & $\sim$ & $>$ & $\sim$ & $\sim$ & $>$ & $\sim$ & $\sim$ & $>$ & $\sim$ & $\sim$ \\
DROP3 & $>$ & $\sim$ & $\sim$ & $\sim$ & $\sim$ & $\sim$ & $\sim$ & $\sim$ & $\sim$ & $\sim$ & $\sim$ & $\sim$ & $\sim$ \\
ENN & $\sim$ & $\sim$ & $\sim$ & $<$ & $\sim$ & $<$ & $\sim$ & $\sim$ & $\sim$ & $\sim$ & $\sim$ & $<$ & $<$ \\
ICF & $>$ & $>$ & $>$ & $\sim$ & $>$ & $\sim$ & $\sim$ & $\sim$ & $\sim$ & $\sim$ & $>$ & $\sim$ & $\sim$ \\
LSBo & $>$ & $>$ & $>$ & $\sim$ & $>$ & $\sim$ & $\sim$ & $\sim$ & $\sim$ & $\sim$ & $>$ & $\sim$ & $\sim$ \\
LSSm & $\sim$ & $\sim$ & $\sim$ & $<$ & $\sim$ & $<$ & $\sim$ & $\sim$ & $<$ & $<$ & $\sim$ & $<$ & $<$ \\
LDIS & $>$ & $>$ & $>$ & $\sim$ & $>$ & $\sim$ & $\sim$ & $>$ & $\sim$ & $\sim$ & $>$ & $\sim$ & $\sim$ \\
ISDSP & $>$ & $>$ & $>$ & $\sim$ & $>$ & $\sim$ & $\sim$ & $>$ & $\sim$ & $\sim$ & $>$ & $\sim$ & $\sim$ \\
\bottomrule
\end{tabular}
\end{center}
\caption{Statistical analysis results with ES for reduction}
\label{t:statResultsES-red}
\end{sidewaystable}

\begin{sidewaystable}
\begin{center}
\setlength{\tabcolsep}{2pt}
\begin{tabular}{llllllllllllll}
\toprule
Algorithm & NoR & NoA & R-0.9 & R-0.1 & SeqP-0.9 & SeqP-0.1 & DROP3 & ENN & ICF & LSBo & LSSm & LDIS & ISDSP \\
\midrule
NoR & $\sim$ & $\sim$ & $\sim$ & $>$ & $\sim$ & $>$ & $>$ & $\sim$ & $>$ & $>$ & $\sim$ & $>$ & $>$ \\
NoA & $\sim$ & $\sim$ & $\sim$ & $>$ & $\sim$ & $>$ & $\sim$ & $\sim$ & $\sim$ & $>$ & $\sim$ & $\sim$ & $>$ \\
R-0.9 & $\sim$ & $\sim$ & $\sim$ & $>$ & $\sim$ & $>$ & $\sim$ & $\sim$ & $>$ & $>$ & $\sim$ & $>$ & $>$ \\
R-0.1 & $<$ & $<$ & $<$ & $\sim$ & $<$ & $\sim$ & $\sim$ & $<$ & $\sim$ & $\sim$ & $<$ & $\sim$ & $\sim$ \\
SeqP-0.9 & $\sim$ & $\sim$ & $\sim$ & $>$ & $\sim$ & $>$ & $\sim$ & $\sim$ & $\sim$ & $>$ & $\sim$ & $>$ & $>$ \\
SeqP-0.1 & $<$ & $<$ & $<$ & $\sim$ & $<$ & $\sim$ & $\sim$ & $<$ & $\sim$ & $\sim$ & $<$ & $\sim$ & $\sim$ \\
DROP3 & $<$ & $\sim$ & $\sim$ & $\sim$ & $\sim$ & $\sim$ & $\sim$ & $\sim$ & $\sim$ & $\sim$ & $<$ & $\sim$ & $\sim$ \\
ENN & $\sim$ & $\sim$ & $\sim$ & $>$ & $\sim$ & $>$ & $\sim$ & $\sim$ & $\sim$ & $>$ & $\sim$ & $>$ & $>$ \\
ICF & $<$ & $\sim$ & $<$ & $\sim$ & $\sim$ & $\sim$ & $\sim$ & $\sim$ & $\sim$ & $\sim$ & $<$ & $\sim$ & $\sim$ \\
LSBo & $<$ & $<$ & $<$ & $\sim$ & $<$ & $\sim$ & $\sim$ & $<$ & $\sim$ & $\sim$ & $<$ & $\sim$ & $\sim$ \\
LSSm & $\sim$ & $\sim$ & $\sim$ & $>$ & $\sim$ & $>$ & $>$ & $\sim$ & $>$ & $>$ & $\sim$ & $>$ & $>$ \\
LDIS & $<$ & $\sim$ & $<$ & $\sim$ & $<$ & $\sim$ & $\sim$ & $<$ & $\sim$ & $\sim$ & $<$ & $\sim$ & $\sim$ \\
ISDSP & $<$ & $<$ & $<$ & $\sim$ & $<$ & $\sim$ & $\sim$ & $<$ & $\sim$ & $\sim$ & $<$ & $\sim$ & $\sim$ \\
\bottomrule
\end{tabular}
\end{center}
\caption{Statistical analysis results with SeqP for accuracy with an SVM classifier}
\label{t:statResultsSeqP-SVM}
\end{sidewaystable}

\begin{sidewaystable}
\begin{center}
\setlength{\tabcolsep}{2pt}
\begin{tabular}{llllllllllllll}
\toprule
Algorithm & NoR & NoA & R-0.9 & R-0.1 & SeqP-0.9 & SeqP-0.1 & DROP3 & ENN & ICF & LSBo & LSSm & LDIS & ISDSP \\
\midrule
NoR & $\sim$ & $\sim$ & $\sim$ & $>$ & $\sim$ & $>$ & $\sim$ & $\sim$ & $>$ & $\sim$ & $\sim$ & $>$ & $>$ \\
NoA & $\sim$ & $\sim$ & $\sim$ & $>$ & $\sim$ & $>$ & $\sim$ & $\sim$ & $>$ & $\sim$ & $\sim$ & $\sim$ & $>$ \\
R-0.9 & $\sim$ & $\sim$ & $\sim$ & $>$ & $\sim$ & $>$ & $\sim$ & $\sim$ & $>$ & $\sim$ & $\sim$ & $>$ & $>$ \\
R-0.1 & $<$ & $<$ & $<$ & $\sim$ & $<$ & $\sim$ & $<$ & $<$ & $\sim$ & $\sim$ & $<$ & $\sim$ & $\sim$ \\
SeqP-0.9 & $\sim$ & $\sim$ & $\sim$ & $>$ & $\sim$ & $>$ & $\sim$ & $\sim$ & $>$ & $\sim$ & $\sim$ & $\sim$ & $>$ \\
SeqP-0.1 & $<$ & $<$ & $<$ & $\sim$ & $<$ & $\sim$ & $\sim$ & $\sim$ & $\sim$ & $\sim$ & $<$ & $\sim$ & $\sim$ \\
DROP3 & $\sim$ & $\sim$ & $\sim$ & $>$ & $\sim$ & $\sim$ & $\sim$ & $\sim$ & $\sim$ & $\sim$ & $\sim$ & $\sim$ & $\sim$ \\
ENN & $\sim$ & $\sim$ & $\sim$ & $>$ & $\sim$ & $\sim$ & $\sim$ & $\sim$ & $\sim$ & $\sim$ & $\sim$ & $\sim$ & $\sim$ \\
ICF & $<$ & $<$ & $<$ & $\sim$ & $<$ & $\sim$ & $\sim$ & $\sim$ & $\sim$ & $\sim$ & $<$ & $\sim$ & $\sim$ \\
LSBo & $\sim$ & $\sim$ & $\sim$ & $\sim$ & $\sim$ & $\sim$ & $\sim$ & $\sim$ & $\sim$ & $\sim$ & $\sim$ & $\sim$ & $\sim$ \\
LSSm & $\sim$ & $\sim$ & $\sim$ & $>$ & $\sim$ & $>$ & $\sim$ & $\sim$ & $>$ & $\sim$ & $\sim$ & $>$ & $>$ \\
LDIS & $<$ & $\sim$ & $<$ & $\sim$ & $\sim$ & $\sim$ & $\sim$ & $\sim$ & $\sim$ & $\sim$ & $<$ & $\sim$ & $\sim$ \\
ISDSP & $<$ & $<$ & $<$ & $\sim$ & $<$ & $\sim$ & $\sim$ & $\sim$ & $\sim$ & $\sim$ & $<$ & $\sim$ & $\sim$ \\
\bottomrule
\end{tabular}
\end{center}
\caption{Statistical analysis results with SeqP for accuracy with a KNN classifier}
\label{t:statResultsSeqP-KNN}
\end{sidewaystable}

\begin{sidewaystable}
\begin{center}
\setlength{\tabcolsep}{2pt}
\begin{tabular}{llllllllllllll}
\toprule
Algorithm & NoR & NoA & R-0.9 & R-0.1 & SeqP-0.9 & SeqP-0.1 & DROP3 & ENN & ICF & LSBo & LSSm & LDIS & ISDSP \\
\midrule
NoR & $\sim$ & $\sim$ & $\sim$ & $<$ & $\sim$ & $<$ & $<$ & $\sim$ & $<$ & $<$ & $\sim$ & $<$ & $<$ \\
NoA & $\sim$ & $\sim$ & $\sim$ & $<$ & $\sim$ & $<$ & $\sim$ & $\sim$ & $<$ & $<$ & $\sim$ & $<$ & $<$ \\
R-0.9 & $\sim$ & $\sim$ & $\sim$ & $<$ & $\sim$ & $<$ & $\sim$ & $\sim$ & $<$ & $<$ & $\sim$ & $<$ & $<$ \\
R-0.1 & $>$ & $>$ & $>$ & $\sim$ & $>$ & $\sim$ & $\sim$ & $>$ & $\sim$ & $\sim$ & $>$ & $\sim$ & $\sim$ \\
SeqP-0.9 & $\sim$ & $\sim$ & $\sim$ & $<$ & $\sim$ & $<$ & $\sim$ & $\sim$ & $<$ & $\sim$ & $\sim$ & $<$ & $<$ \\
SeqP-0.1 & $>$ & $>$ & $>$ & $\sim$ & $>$ & $\sim$ & $\sim$ & $>$ & $\sim$ & $\sim$ & $>$ & $\sim$ & $\sim$ \\
DROP3 & $>$ & $\sim$ & $\sim$ & $\sim$ & $\sim$ & $\sim$ & $\sim$ & $\sim$ & $\sim$ & $\sim$ & $>$ & $\sim$ & $\sim$ \\
ENN & $\sim$ & $\sim$ & $\sim$ & $<$ & $\sim$ & $<$ & $\sim$ & $\sim$ & $<$ & $\sim$ & $\sim$ & $<$ & $<$ \\
ICF & $>$ & $>$ & $>$ & $\sim$ & $>$ & $\sim$ & $\sim$ & $>$ & $\sim$ & $\sim$ & $>$ & $\sim$ & $\sim$ \\
LSBo & $>$ & $>$ & $>$ & $\sim$ & $\sim$ & $\sim$ & $\sim$ & $\sim$ & $\sim$ & $\sim$ & $>$ & $\sim$ & $\sim$ \\
LSSm & $\sim$ & $\sim$ & $\sim$ & $<$ & $\sim$ & $<$ & $<$ & $\sim$ & $<$ & $<$ & $\sim$ & $<$ & $<$ \\
LDIS & $>$ & $>$ & $>$ & $\sim$ & $>$ & $\sim$ & $\sim$ & $>$ & $\sim$ & $\sim$ & $>$ & $\sim$ & $\sim$ \\
ISDSP & $>$ & $>$ & $>$ & $\sim$ & $>$ & $\sim$ & $\sim$ & $>$ & $\sim$ & $\sim$ & $>$ & $\sim$ & $\sim$ \\
\bottomrule
\end{tabular}
\end{center}
\caption{Statistical analysis results with SeqP for reduction}
\label{t:statResultsSeqP-red}
\end{sidewaystable}

\end{document}